\documentclass{article}

\usepackage{arxiv}
\usepackage{mfirstuc}
\usepackage[utf8]{inputenc} 
\usepackage[T1]{fontenc}    
\usepackage{hyperref}       
\usepackage{url}            
\usepackage{booktabs}       
\usepackage{amsfonts}       
\usepackage{nicefrac}       
\usepackage{microtype}      
\usepackage{tabularx}
\usepackage{ragged2e}
\usepackage{lipsum}		
\usepackage{graphicx}
\usepackage{natbib}
\usepackage{doi}
\usepackage{amsmath}
\usepackage{threeparttable}
\usepackage{subcaption}
\usepackage{algorithm}
\usepackage{algpseudocode}

\title{Landslide Hazard Mapping with Geospatial Foundation Models: Geographical Generalizability, Data Scarcity, and Band Adaptability}

\author{ 
	\href{https://orcid.org/0000-0000-0000-0000}{\includegraphics[scale=0.06]{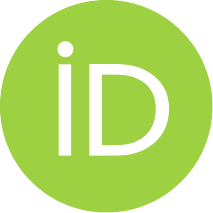}\hspace{1mm}Wenwen Li}\thanks{Corresponding author} \\
	School of Geographical Sciences and Urban Planning\\
	Arizona State University\\
	Tempe, AZ 85287 \\
	\texttt{wenwen@asu.edu} \\
	\And
	\href{https://orcid.org/0000-0000-0000-0000}{\includegraphics[scale=0.06]{orcid.pdf}\hspace{1mm}Sizhe Wang} \\
    School of Geographical Sciences and Urban Planning\\
    School of Computing and Augmented Intelligence\\
    Arizona State University\\
	Tempe, AZ 85287 \\
    \texttt{wsizhe@asu.edu} \\
	\And
	\href{https://orcid.org/0000-0000-0000-0000}{\includegraphics[scale=0.06]{orcid.pdf}\hspace{1mm}Hyunho Lee} \\
    School of Geographical Sciences and Urban Planning\\
	Arizona State University\\
	Tempe, AZ 85287 \\
	\texttt{hlee401@asu.edu} \\
    \And
	\href{https://orcid.org/0000-0000-0000-0000}{\includegraphics[scale=0.06]{orcid.pdf}\hspace{1mm}Chenyan Lu} \\
    School of Geographical Sciences and Urban Planning\\
	Arizona State University\\
	Tempe, AZ 85287 \\
	\texttt{chenya23@asu.edu} \\
	\And
	\href{https://orcid.org/0000-0000-0000-0000}{\includegraphics[scale=0.06]{orcid.pdf}\hspace{1mm}Sujit Roy} \\
    NASA Marshall Space Flight Center\\
	Huntsville, AL, USA \\
	\texttt{sujit.roy@uah.edu} \\
    	\And
	\href{https://orcid.org/0000-0000-0000-0000}{\includegraphics[scale=0.06]{orcid.pdf}\hspace{1mm}Rahul Ramachandran} \\
    NASA Marshall Space Flight Center\\
	Huntsville, AL, USA \\
	\texttt{rahul.ramachandran@nasa.gov} \\
	\And
	\href{https://orcid.org/0000-0000-0000-0000}{\includegraphics[scale=0.06]{orcid.pdf}\hspace{1mm}Chia-Yu Hsu} \\
    School of Geographical Sciences and Urban Planning\\
	Arizona State University\\
	Tempe, AZ 85287 \\
	\texttt{chsu53@asu.edu} \\
}

\renewcommand{\shorttitle}{}

\begin{document}
\maketitle

\begin{abstract}
Landslides cause severe damage to lives, infrastructure, and the environment, making accurate and timely mapping essential for disaster preparedness and response. However, conventional deep learning models often struggle when applied across different sensors, regions, or under conditions of limited training data. To address these challenges, we present a three-axis analytical framework of sensor, label, and domain for adapting geospatial foundation models (GeoFMs), focusing on Prithvi-EO-2.0 for landslide mapping. Through a series of experiments, we show that it consistently outperforms task-specific CNNs (U-Net, U-Net++), vision transformers (Segformer, SwinV2-B), and other GeoFMs (TerraMind, SatMAE). The model, built on global pretraining, self-supervision, and adaptable fine-tuning, proved resilient to spectral variation, maintained accuracy under label scarcity, and generalized more reliably across diverse datasets and geographic settings. Alongside these strengths, we also highlight remaining challenges such as computational cost and the limited availability of reusable AI-ready training data for landslide research. Overall, our study positions GeoFMs as a step toward more robust and scalable approaches for landslide risk reduction and environmental monitoring.
\end{abstract}


\begin{figure}[ht]
	\centering
	\includegraphics[width=\textwidth]{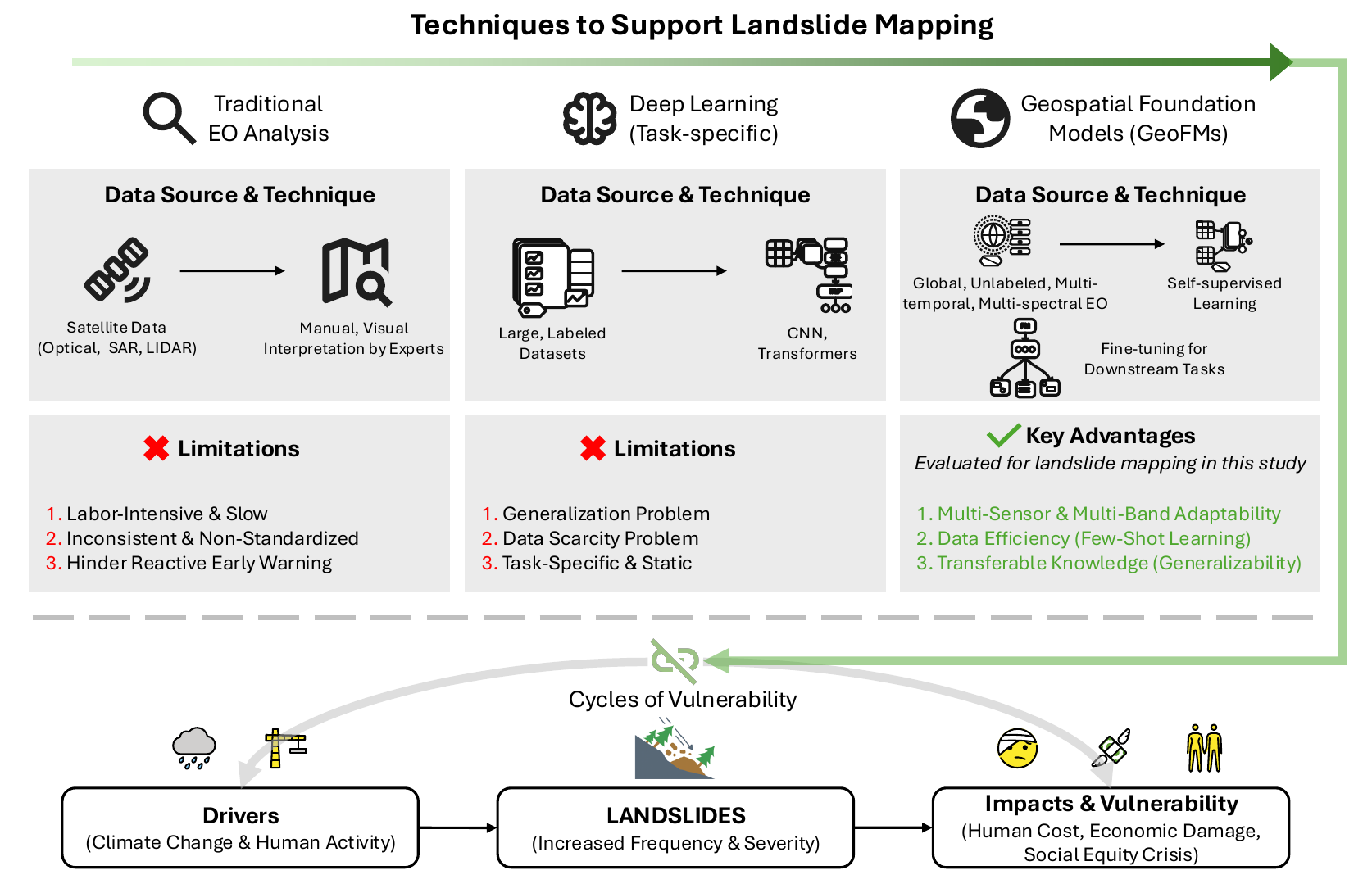}
	\caption{Conceptual framework illustrating the evolution of landslide mapping techniques and their role in breaking the cycle of vulnerability. Top panel: transition from traditional expert-driven analysis, to task-specific deep learning, to geospatial foundation models (GeoFMs) that leverage self-supervised learning and fine-tuning for downstream tasks. The green arrow represents the technological shift toward scalable, adaptable, and data-efficient solutions. Bottom panel: the cycle of vulnerability, where climate change and human activities drive landslides, leading to impacts that increase community vulnerability. By enabling multi-band adaptability, few-shot capability, and transferable knowledge, GeoFMs provide a pathway to proactive risk management and aim to disrupt this reinforcing cycle. \protect\footnotemark}
	\label{fig:overview}
\end{figure}

\section{Introduction}

\footnotetext{All icons/emojis are from Material Symbols \& Icons (Apache License 2.0) or OpenMoji (CC BY-SA 4.0).}

Landslides are among the most widespread and destructive natural hazards, posing significant threats to lives, infrastructure, and ecosystems \citep{who_landslides}. Between 1998 and 2017, landslides affected an estimated 4.8 million people and caused more than 18,000 deaths \citep{who_landslides}. They accounted for 17\% of all fatalities from natural hazards between 2004 and 2016 \citep{li2022strategic}. Landslide risk is intensifying as its primary drivers, climate change and human activity, continue to accelerate. The Intergovernmental Panel on Climate Change (IPCC) warns that more frequent and intense rainfall will likely trigger additional landslides \citep{seneviratne2012changes}. Human activities such as deforestation and road construction in mountainous regions weaken slopes, increasing the likelihood of landslides and leaving communities more vulnerable \citep{li2022strategic}.

Beyond immediate deaths, landslides lead to a chain of destructive outcomes, including severe injuries, long-term psychological distress, and the destruction of infrastructure. This damage can delay emergency response and create public health crises \citep{eea_landslides}. These impacts, which include displacement, livelihood loss, and billions of dollars in annual damages, are not isolated events. Instead, they trap communities in a reinforcing cycle of vulnerability, as illustrated in the bottom panel of Figure~\ref{fig:overview}. The destruction from one landslide reduces a community's ability to recover, which leaves it more exposed and less resilient to the next one \citep{li2022strategic}. This harmful cycle disproportionately affects the world's most vulnerable populations, who often live in high-risk areas and lack the resources to rebuild. As a result, landslide risk has become an urgent issue of social equity. Breaking this cycle is essential for sustainable development in high-risk regions. This requires effective, large-scale systems for landslide monitoring, mapping, and risk assessment to build resilience before a disaster occurs.

To meet this need, Earth Observation (EO) technologies are increasingly used to monitor and map landslides, as the vast and hazardous nature of landslide-prone terrain makes large-scale field surveys impractical \citep{scaioni2014remote, zeng2024can}. Remote sensing techniques, such as optical imagery, Synthetic Aperture Radar (SAR), and LiDAR, have greatly advanced landslide research \citep{mondini2021landslide, eeckhaut2007use, scaioni2014remote}. However, the conventional analysis approach faces critical bottlenecks. Landslide inventories, which form the foundation of hazard assessment, are still predominantly generated through manual visual interpretation by experts \citep{guzzetti2012landslide}. This process is labor-intensive, slow, and inconsistent, limiting the development of standardized risk assessments \citep{galli2008comparing}. More importantly, it remains reactive: inventories are often produced too late to support effective early warning and rapid response. Addressing the gap between the urgent need for proactive risk management and the limitations of traditional EO methods requires a fundamental shift in approach, a direction this paper explores and the top panel of Figure~\ref{fig:overview} illustrates.

The first major step in this direction came with the application of deep learning (DL). Models such as Convolutional Neural Networks (CNNs) were trained to automatically recognize landslide patterns in satellite imagery, offering a level of automation and objectivity that manual methods lacked \citep{yang2024exploring, zhang2024deep}. While a significant advance, this first generation of task-specific deep learning models introduced its own set of limitations.

First is the generalization problem. It limits the development of transferable knowledge. Standard deep learning models are trained to recognize patterns within a given dataset, but the appearance of landslides varies widely with local geology, vegetation, and triggering mechanism. A model trained on shallow debris flows in a tropical forest, for example, might fail when applied to deep-seated rotational slides in a semi-arid region \citep{ghorbanzadeh2021comprehensive, soares2022landslide}. This ``weak replicability'' means that knowledge is not easily transferred, requiring new models to be trained for different geographic areas \citep{goodchild2021replication}.

Second is the data scarcity problem. It limits data efficiency and rapid response. This challenge is twofold. One, because landslides are infrequent events, collecting the large number of distinct examples required by ``data-hungry'' deep learning models is often scientifically and economically untenable. Two, individual satellite images exhibit severe class imbalance, where the landslide itself covers only a fraction of the terrain compared to stable ground \citep{chen2022deep, yu2024semantic}. This dual bottleneck of scarce events and imbalanced data creates a strong dependency on large, pre-existing datasets. Such a requirement hinders the rapid deployment of models in data-sparse regions or in the immediate aftermath of disasters \citep{shyalika2024comprehensive}. As a result, task-specific models often remain too static and narrowly focused to address problems at a global scale.

In response to these limitations, a new modeling approach, Geospatial foundation models (GeoFMs), has emerged. Instead of training a small, specialized model for a single task, GeoFMs are pre-trained on vast, unlabeled, multi-spectral satellite imagery using self-supervised learning techniques that do not require manual labels \citep{manas2021seasonal, cong2022satmae}. For example, the Prithvi-EO-2.0 model was pre-trained on millions of image sequences from NASA's global Harmonized Landsat and Sentinel-2 (HLS) archive \citep{szwarcman2024prithvi}. This large-scale pre-training enables the model to learn rich, general-purpose representations of Earth's surface. The resulting model is a versatile foundational model that can be rapidly adapted, or ``fine-tuned'', for landslide detection and other downstream tasks with minimal labeled data.

The core advantages of GeoFMs directly address the shortcomings of previous methods. Their native ability to process multi-sensor and multi-spectral data provides critical adaptability for modeling dynamic processes under varying data availability \citep{hsu2024geospatial, jakubik2025terramind}. They also enhance data efficiency by enabling fine-tuning with only a small number of labeled examples, allowing for rapid deployment in ongoing hazard emergencies \citep{jakubik2023foundation}. Finally, by learning from diverse global datasets, GeoFMs promote transferable knowledge, making them more generalizable across different environments \citep{li2023assessment}.

Therefore, this study systematically evaluates these three capabilities in the context of landslide mapping. We test the performance of the Prithvi-EO-2.0 model, framing these technical assessments as solutions to the core scientific challenges and as a means to help break the cycle of vulnerability. Specifically, this paper addresses the following research questions:

\begin{itemize}

	\item How does Prithvi-EO-2.0 perform in a standard landslide mapping task compared to task-specific deep learning models and other GeoFMs?
	
	\item How adaptable is the model to varied data inputs? We evaluate its performance when fine-tuned with different spectral band combinations to reflect real-world scenarios where sensor availability varies.
	
	\item What is the model's potential for rapid response in data-scarce environments? We quantify its performance when fine-tuned with a reduced number of local training examples.
	
	\item To what extent can the model generalize and handle domain shifts? We evaluate this by training the model on one geographic region or dataset and testing its ability to detect landslides in other regions and/or datasets.
	
\end{itemize}

The remainder of the paper is organized as follows. Section 2 reviews the literature on data sources and AI methods applied to landslide studies. Section 3 introduces the Prithvi-EO foundation model and outlines our strategies for evaluating and adapting it to landslide mapping. Section 4 presents the experimental results. Section 5 discusses the remaining challenges of leveraging geospatial foundation models for landslide research. Finally, Section 6 concludes the paper and proposes future directions.

\section{Related Work} \label{sec:literature}
\subsection{Foundations of Landslide Inventorying}

Traditionally, landslide mapping has relied on geomorphological field surveys and the visual interpretation of stereoscopic aerial photographs \citep{scaioni2014remote}. These methods, long considered the \textit{gold standard}, offer high contextual fidelity by leveraging morphological signatures such as head scarps, hummocky deposits, and flow tracks. Historical aerial photo archives have been invaluable for creating multi-temporal inventories and tracking slope failure dynamics \citep{cardinali2002geomorphological, guzzetti2006landslide}. Yet, despite their accuracy, these methods are time-consuming, labor-intensive, and inherently subjective, with quality depending heavily on the interpreter's expertise \citep{guzzetti2012landslide}. As a result, large-scale or rapid-response mapping remains impractical, leaving vast landslide-prone areas unmapped.

To overcome these scalability issues, satellite-based Earth Observation (EO) technologies initiated a major advance in landslide studies by enabling systematic, large-scale monitoring of terrain dynamics. Among these, optical imagery from missions such as Landsat provided the synoptic views of Earth's surface, facilitating landslide mapping through the spectral contrast between exposed earth and vegetated surfaces \citep{novellino2024mapping, wen2022landslide}. However, optical approaches are limited by cloud interference \citep{handwerger2022generating}. Synthetic Aperture Radar (SAR) offered an advantage by overcoming the constraint through active microwave sensing and enabling observations regardless of lighting or weather conditions \citep{massonnet1998radar}. Light Detection and Ranging (LiDAR) has further advanced the field, especially in densely forested regions. By penetrating dense canopies, LiDAR generates high-resolution digital elevation models of the bare ground and helps reveal subtle topographic signatures of both recent and ancient landslides previously obscured by vegetation \citep{mckean2004objective, jaboyedoff2012use}.

While these EO technologies improved data acquisition, they introduced new challenges in analysis and interpretation. The diversity of sensor outputs fractured the morphology-based definition of a landslide into multiple, sensor-dependent manifestations, e.g., a spectral anomaly in optical imagery or a fine-scale topographic imprint in LiDAR. These definitions are complementary but not interchangeable \citep{casagli2017spaceborne}. For example, a sudden debris flow is obvious in optical imagery but may be too chaotic for radar interferometry. This fragmentation highlighted the need for integrative approaches capable of fusing heterogeneous data streams. It also laid the groundwork for the next major shift that apply artificial intelligence to automate and synthesize multimodal observations.

\subsection{Task-Specific Deep Learning}

The second major shift in landslide mapping was driven by deep learning (DL), which shifted the methodology from human-defined rules to data-driven pattern recognition \citep{zhang2024deep, chen2025harnessing}. The initial breakthrough came with Convolutional Neural Networks (CNNs). At their core, CNNs employ stacked layers of convolutional filters that automatically learn a hierarchy of features from raw pixel data. Early layers learn to detect simple patterns like edges and textures, while deeper layers combine these to recognize more complex morphological features, such as head scarps or debris deposits. This ability to learn relevant features directly from imagery, rather than relying on hand-crafted rules, allowed CNNs to outperform prior machine learning approaches in landslide detection tasks \citep{oak2024comparative, chen2025harnessing}.

For landslide mapping, the primary task evolved from simple image-level categorization to semantic segmentation which performs the pixel-level classification of an entire image. A cornerstone architecture for this task is the U-Net, a specialized type of CNN \citep{ronneberger2015u, chen2023landslide}. It features a symmetric encoder-decoder structure; the encoder path progressively downsamples the image to capture broad contextual information, while the decoder path upsamples this information to reconstruct a high-resolution segmentation map. The key innovation of U-Net lies in its ``skip connections,'' which feed feature maps from the encoder directly to corresponding layers in the decoder, preserving fine-grained spatial details essential for delineating precise landslide boundaries. More recently, transformer-based architectures, such as the Swin Transformer \citep{liu2021swin}, have introduced a new approach. Originally designed for natural language, Swin Transformers use a self-attention mechanism to weigh the importance of all pixels in an image relative to each other. This enables the model to capture long-range spatial dependencies and global context, an advantage for distinguishing complex landslide features from spectrally similar landforms like quarries or riverbeds \citep{syed2024semantic, rs16234464}.

Despite strong localized performance, task-specific DL models face two limitations that undermine their scalability. First, generalization remains a challenging task for existing deep learning models. This is because the visual and morphological expression of landslides varies with geology, vegetation, climate, and triggering mechanisms, so models tuned to one setting (e.g., shallow debris flows in humid tropics) often degrade when deployed elsewhere (e.g., deep-seated slides in semi-arid regions) \citep{ganerod2023globally, prakash2021new}. Second, data scarcity persists. DL models are inherently data-hungry, requiring extensive annotated datasets. Producing landslide inventories at this scale is labor-intensive \citep{xu2024cas}. The problem is further complicated by class imbalance, as landslide pixels are exceedingly rare compared to stable terrain \citep{yu2024semantic}. Mitigation strategies such as augmentation, re-weighted losses, or focal loss \citep{lin2017focal} offer incremental improvements but do not eliminate the core dependency on abundant, high-quality labeled data. 

These two challenges are tightly coupled in a self-reinforcing cycle. A model's failure to generalize to new regions demands more labeled data, but the cost of producing such data means the global training pool remains sparse and fragmented. Models trained on this limited data inevitably inherit poor generalization capabilities, perpetuating the cycle.

This challenge points to a limitation of task-specific approaches. A promising way forward is to develop methods that learn broad, foundational representations from vast, unlabeled global data and then fine-tune them for specific landslide mapping tasks.

\subsection{The Geospatial Foundation Model Approach}

In response to the limitations of task-specific models, a transformative new approach has emerged: the foundation model. Originating in Natural Language Processing (NLP) with landmark systems such as GPT, and later adapted for computer vision with influential models such as DINOv2 \citep{oquab2023dinov2}, foundation models represent a fundamental shift in AI design. Instead of developing a model from scratch for each task, a single model undergoes large-scale pre-training on diverse, unlabeled datasets. The cornerstone of this process is Self-Supervised Learning (SSL), in which the model learns general-purpose representations by solving proxy tasks, such as reconstructing masked portions of the input, without requiring manual labels \citep{he2022masked}. The resulting model serves as a powerful ``foundation'' that can be fine-tuned efficiently for downstream applications with minimal labeled data.

The adaptation of this concept to Earth Observation has given rise to GeoFMs. These models are not simply vision architectures repurposed for satellite imagery; they are designed to handle the unique challenges of satellite data. GeoFMs ingest information across multiple spectral bands beyond the visible, account for multi-temporal dynamics that encode seasonal cycles and event-driven change, and exploit spatiotemporal context, including location and time of acquisition, and other metadata, that strongly conditions surface appearance \citep{xiao2025foundation,szwarcman2024prithvi}. 

Early milestones include SatMAE, which adapted the Masked Autoencoder (MAE) framework for EO data. Its self-supervised task involved reconstructing masked 3D spatio-temporal-spectral patches from image time series, forcing the model to learn the intrinsic relationships between space, time, and spectral bands \citep{cong2022satmae}. Building on this foundation, Prithvi-EO-2.0 has set a new benchmark by enhancing the model's contextual understanding. While also pre-trained on 3D spatiotemporal data from NASA's HLS archive, its key innovation is the explicit encoding of date and location metadata directly into the model's input embeddings. This provides deep context, enabling the model to account for geographic and seasonal variance in a way that prior methods could not \citep{szwarcman2024prithvi}. More recently, models such as Terramind have advanced this approach toward true multimodality. To fuse heterogeneous inputs, they typically employ a multi-encoder architecture where separate pathways process optical, SAR, and DEM data, with the outputs projected into a shared representation space to construct a holistic, multi-sensor understanding of Earth systems \citep{jakubik2025terramind}.

This shift, from treating satellite imagery as generic pictures to modeling it as a geospatial knowledge system, is critical. By capturing global, multimodal, and context-rich representations, GeoFMs address the generalization and data scarcity bottlenecks that limit task-specific models.

\subsection{Bridging the Research Gap: Adopting GeoFMs for Landslide Mapping}

GeoFMs directly address the core challenges of generalization and data scarcity that have constrained task-specific models in landslide mapping \citep{jakubik2023foundation}. By learning from vast, unlabeled, and globally diverse satellite data, GeoFMs learn a foundational understanding of Earth's surface processes. This pre-trained knowledge is expected to make them more transferable across diverse geographic settings \citep{li2023assessment} and more robust in few-shot learning scenarios, which are critical for time-sensitive disaster response scenarios \citep{szwarcman2024prithvi}.

However, research applying GeoFMs to landslide mapping remains largely unexplored. Early studies have primarily adapted general-purpose Vision Foundation Models (VFMs), such as the Segment Anything Model (SAM), through lightweight fine-tuning strategies (e.g., TransLandSeg, LandslideNet) \citep{hou2024translandseg, yu2024landslidenet, kirillov2023segment}. While these approaches achieve competitive results, they remain inherently constrained: VFMs are trained on natural RGB imagery and lack exposure to multi-spectral, multi-temporal, and metadata-rich EO data. As a result, VFM-based approaches demonstrate feasibility but fail to fully exploit the rich physical information and contextual dependencies captured by geospatial-native pretraining. 

In this work, we present the first systematic evaluation and adaptation of a GeoFM, Prithvi-EO-2.0, for landslide mapping. Unlike VFM-based methods, we leverage its capacity to integrate multi-spectral inputs to achieve improved generalization and robust few-shot performance. This study establishes a comprehensive benchmark for GeoFMs in landslide science, providing a physically grounded and scalable framework for global landslide mapping.

\begin{table}[ht]
	\begin{threeparttable}
	\begingroup
	\renewcommand{\arraystretch}{1.25}
	\caption{Scientific relevance of the Landslide4Sense data layers for landslide mapping. The integration of multi-spectral Sentinel-2 bands and ALOS PALSAR topography provides a physically grounded foundation for evaluating Prithvi-EO-2.0's adaptability to diverse sensing configurations.}
	\centering
	\begin{tabularx}{\textwidth}{>{\raggedright}p{3.3cm} >{\RaggedRight}p{4cm} >{\RaggedRight}X}
		\toprule
		\textbf{Data Type} & \textbf{Source / Bands} & \textbf{Scientific Relevance for Landslide Detection} \\
		\midrule
		Atmospheric / aerosol & Sentinel-2 Band 1 & Provides information on atmospheric haze and scattering, contributing to surface reflectance interpretation. \\
		
		Visible spectrum \newline (Blue, Green, Red) & Sentinel-2 Bands 2, 3, 4 & Captures natural-color imagery, useful for detecting exposed soil, fresh scars, and vegetation contrasts when delineating landslide boundaries. \\
		
		Red-edge bands & Sentinel-2 Bands 5, 6, 7 & Capture subtle vegetation changes (chlorophyll, canopy structure), enhancing the detection of slope disturbances that may be missed by standard RGB or NIR. \\
		
		Near-infrared \newline (NIR) & Sentinel-2 Band 8\tnote{*} & Sensitive to vegetation cover and canopy density. Drops in NIR signal vegetation loss, aiding landslide identification and vegetation index computation (e.g., NDVI). \\
		
		Water vapor & Sentinel-2 Band 9 & Captures atmospheric water vapor, supporting correction of surface reflectance and potential integration with environmental conditions. \\
		
		Cirrus detection & Sentinel-2 Band 10 & Designed for cirrus cloud identification, supporting quality control of optical imagery. \\
		
		Short-wave infrared \newline (SWIR) & Sentinel-2 Bands 11, 12 & Sensitive to soil moisture and mineral content. Helps detect bare-earth surfaces and wet, unstable slopes prone to landslides. \\
		
		Topographic features & ALOS PALSAR (DEM + derived slope) & Provides terrain attributes. Steep slopes and concave topography are key predictors of gravitational instability. \\
		\bottomrule
	\end{tabularx}
	\begin{tablenotes}
		\item[*] Landslide4Sense includes Sentinel-2 Band 8 (NIR broad) rather than Band 8A (NIR narrow), which was used during Prithvi-EO-2.0 pretraining. 
		\item[\textdagger] Resolution note: All spectral bands (including native 20\,m and 60\,m) and the topographic layers are resampled to 10\,m/pixel in the released dataset.
	\end{tablenotes}
	\label{tab:bands}
	\endgroup
	\end{threeparttable}
\end{table}

\section{Methodology}
\subsection{Study Regions and Landslide Inventory Datasets} \label{sec:dataset}

\begin{figure}[ht]
	\centering
	\begin{subfigure}{0.47\textwidth}
	\includegraphics[width=\linewidth]{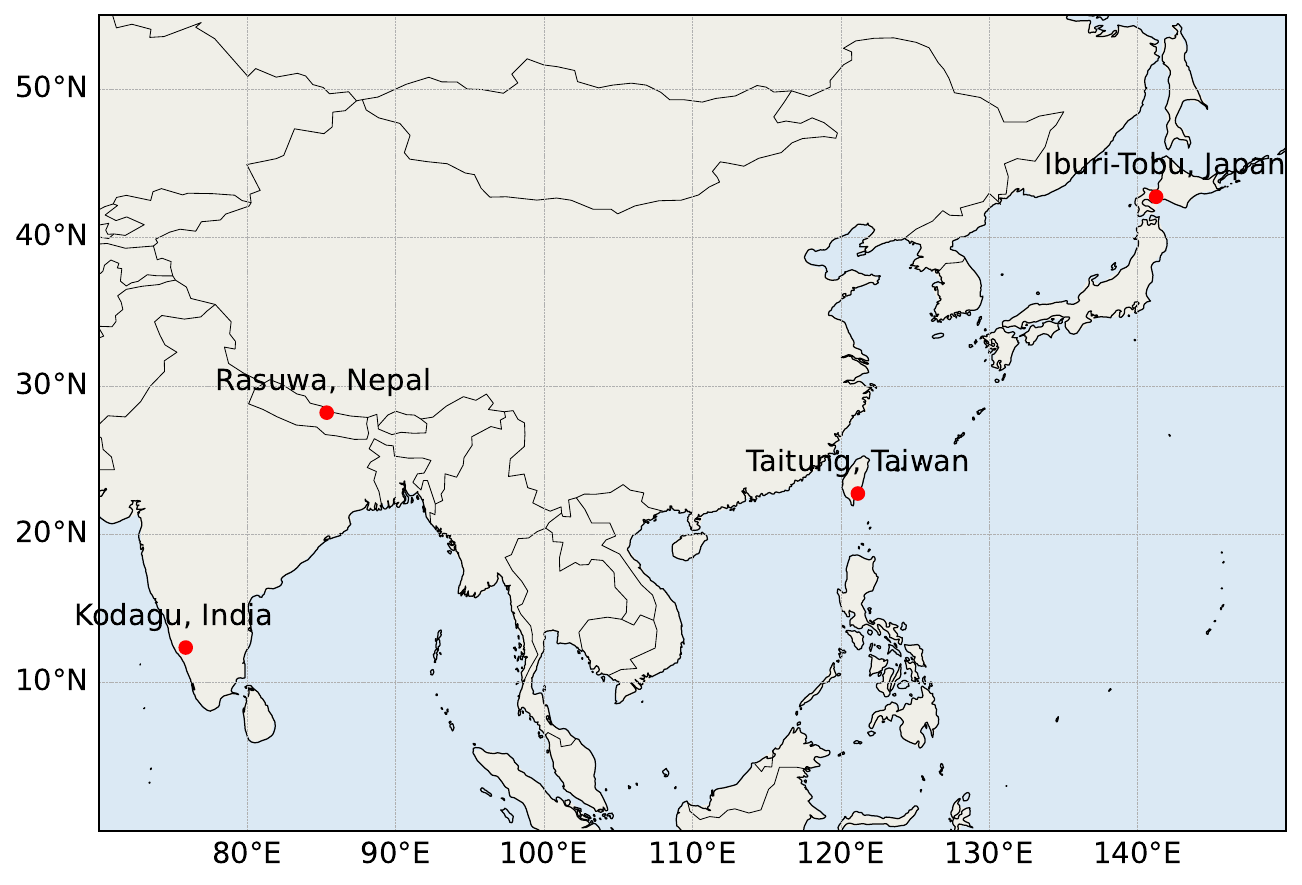}	
	\caption{} \label{fig:dataset_geo_loc}
	\end{subfigure}
	\begin{subfigure}{0.52\textwidth}
	\includegraphics[width=\linewidth]{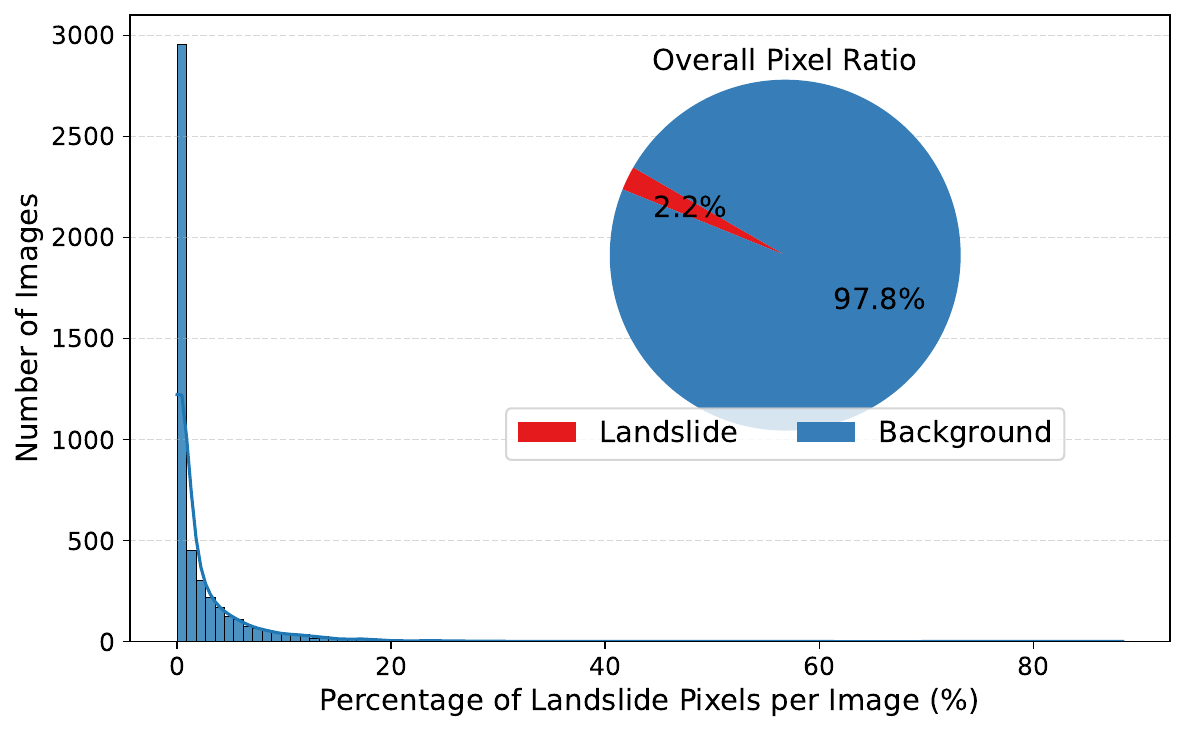}
	\caption{} \label{fig:dataset_class_imbalance}
	\end{subfigure}
	\begin{subfigure}{\textwidth}
	\includegraphics[width=\linewidth]{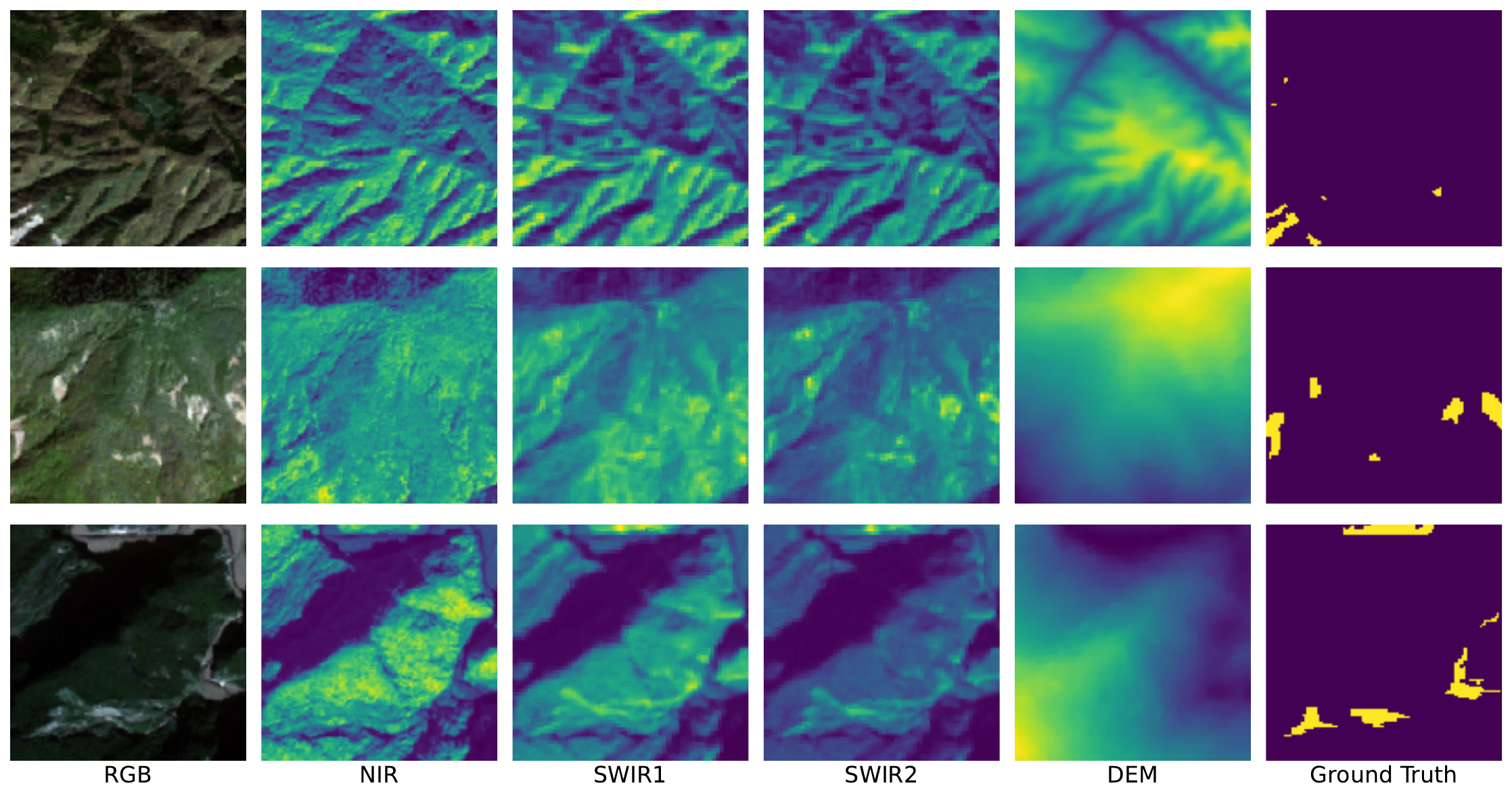}
	\caption{} \label{fig:dataset_multi_sensor}
	\end{subfigure}
	\caption{Overview of the Landslide4Sense benchmark dataset. 
	(a) Study regions spanning diverse climatic, geological, and geomorphic settings. These sites capture landslide activity triggered by both earthquakes and extreme rainfall. 
	(b) Class distribution of landslide versus background pixels, highlighting the extreme class imbalance: landslides constitute only 2.2\% of the total labeled area, and most image patches contain fewer than 10\% landslide pixels. 
	(c) Example multi-sensor inputs, including optical (RGB), infrared (NIR, SWIR), and topographic (DEM) data, alongside the ground truth mask. In the released dataset, all channels, including native 20\,m/60\,m Sentinel-2 bands and the topographic layers, are resampled to 10\,m/pixel.}
	\label{fig:dataset_overview}
\end{figure}

To evaluate the Prithvi-EO-2.0 foundation model, we use the Landslide4Sense dataset, a publicly available, multi-sensor benchmark for semantic landslide segmentation \citep{ghorbanzadeh2022outcome, ghorbanzadeh2022landslide4sense}. The dataset contains 4,844 image patches of size $128 \times 128$ pixels with pixel-level ground truth masks, following a predefined split into training (3,799), validation (245), and test (800) sets to ensure benchmarking consistency. In the released dataset, all channels, including Sentinel-2 bands with native 20\,m/60\,m resolution and the topographic layers, are resampled to 10\,m/pixel to provide a uniform interface. An overview of the study regions, class distribution, and input modalities is provided in Figure~\ref{fig:dataset_overview}. 

The dataset spans four regions across Asia, each with distinct landslide triggers and settings (Figure~\ref{fig:dataset_geo_loc}). It covers earthquake-induced failures in Japan and Nepal as well as rainfall-driven slides in Taiwan and India, representing a spectrum of landslide expressions typical of these trigger mechanisms. This diversity allows us to evaluate whether Prithvi-EO-2.0 can capture generalizable physical signatures of landslides rather than overfitting to site-specific visual patterns.

Each patch integrates 14 data channels from Sentinel-2 spectral bands and ALOS PALSAR topography, providing a rich multi-modal representation of terrain conditions (Table~\ref{tab:bands}; Figure~\ref{fig:dataset_multi_sensor}). This combination enables both realistic monitoring scenarios and controlled experiments on sensor adaptability. In particular, we evaluate the model under multiple input configurations that range from full multi-sensor fusion to reduced spectral subsets. These include a configuration designed to approximate Prithvi's six-band pretraining interface (B2, B3, B4, B8A, B11, B12). Because Landslide4Sense provides B8 (NIR broad) rather than B8A (NIR narrow), we substitute B8 for B8A in this configuration. By varying the available inputs, we test how robustly the model adapts to different sensing conditions without depending on a fixed set of bands.

However, the dataset also introduces a significant class-imbalance challenge. As shown in Figure~\ref{fig:dataset_class_imbalance}, landslide pixels constitute only 2.2\% of the total labeled area, with most patches containing less than 10\% landslide coverage. This severe imbalance mirrors real-world conditions but complicates both training and evaluation, requiring careful consideration of multiple performance metrics to assess model robustness under highly skewed distributions. 

\subsection{The Prithvi-EO-2.0 Model: A Geospatial-Native Foundation Model} \label{sec:prithvi_model}

\begin{figure}[ht]
	\centering
	\includegraphics[width=\linewidth]{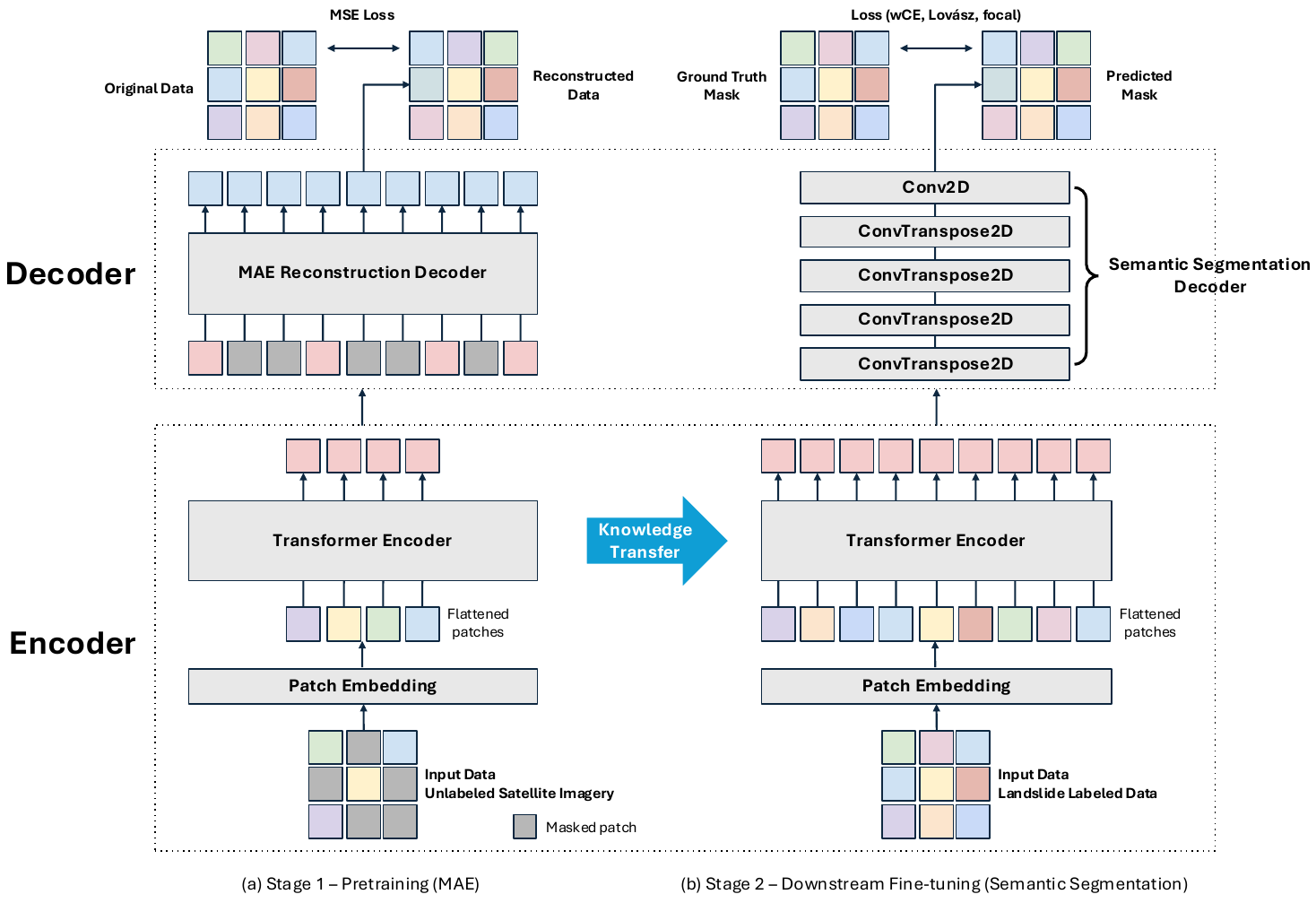}
	\caption{Overview of the Prithvi-EO-2.0 architecture and fine-tuning framework. 
	(a) Pretraining stage (MAE): Multi-band EO imagery is divided into non-overlapping patches, and a subset is masked. A transformer encoder with an MAE reconstruction decoder is trained to reconstruct the missing patches by minimizing mean squared error (MSE). 
	(b) Downstream fine-tuning stage (semantic segmentation): The pretrained encoder is paired with a lightweight convolutional decoder (Conv2D and ConvTranspose2D layers) to produce per-pixel landslide segmentation masks. The decoder is trained with imbalance-aware loss functions, including weighted cross-entropy (wCE), Lovász loss, and focal loss.}
	\label{fig:prithvi_architecture}
\end{figure}

In this study, we evaluate Prithvi-EO-2.0, a geospatial foundation model (GeoFM) jointly developed by IBM and NASA \citep{szwarcman2024prithvi}, to examine whether its large-scale pretraining confers the properties needed for robust landslide mapping: efficient feature reuse, improved generalization, and adaptability to diverse sensing configurations. Prithvi-EO-2.0 is built on a Vision Transformer (ViT) backbone and pretrained with a Masked Autoencoder (MAE) strategy on 4.2 million multi-temporal image sequences from NASA's Harmonized Landsat-Sentinel-2 (HLS) archive at 30\,m resolution \citep{dosovitskiy2020image,he2022masked,claverie2018harmonized,ju2025harmonized}. The pretraining used six HLS spectral bands (Blue-B2, Green-B3, Red-B4, NIR narrow-B8A, SWIR1-B11, SWIR2-B12), providing sensitivity to vegetation, soil moisture, and surface mineralogy. During pretraining (Figure~\ref{fig:prithvi_architecture}a), input patches are partially masked, and the model learns to reconstruct the missing regions by minimizing mean squared error (MSE). This self-supervised task enables the encoder to build spatio-temporal representations that integrate spectral, morphological, and contextual information from globally diverse data.

We systematically assess two model variants, Prithvi-EO-2.0-300M and Prithvi-EO-2.0-600M, which differ in parameter scale (300M vs.\ 600M). This comparison allows us to probe how model capacity influences performance, data efficiency, and cross-region transferability. 

For downstream semantic segmentation of landslides, we adopt a two-stage framework consisting of a pretrained Prithvi encoder $F_{\theta}$ and a lightweight convolutional decoder $G_{\phi}$ composed of Conv2D and ConvTranspose2D layers (Figure~\ref{fig:prithvi_architecture}b). Because Landslide4Sense provides single-date imagery, we fine-tune with $T=1$; the model's temporal embeddings remain part of the encoder but receive a single timestamp during training. 

Let $\mathbf{X}\in\mathbb{R}^{H\times W\times B}$ denote a multi-band Earth observation (EO) patch, where $B$ is the number of spectral channels (default $B=6$ to match Prithvi's pretraining interface, though other configurations are tested in Section~\ref{sec:spectral_band_combinations}). The encoder maps $\mathbf{X}$ to latent features $\mathbf{Z}$:
\[
\mathbf{Z} = F_{\theta}(\mathbf{X}) \, .
\]
The decoder then upsamples $\mathbf{Z}$ to produce per-pixel logits $\mathbf{S}\in\mathbb{R}^{H\times W\times C}$, with $C=2$ for landslide and non-landslide classes:
\[
\mathbf{S} = G_{\phi}(\mathbf{Z}) \, .
\]
Applying a softmax activation yields per-pixel probabilities, from which the final segmentation mask is derived. 
\[
\hat{\mathbf{Y}}=\sigma(\mathbf{S}) \quad \text{($\sigma$ is softmax over } C=2 \text{ classes).} 
\]

Because the Landslide4Sense dataset exhibits extreme class imbalance, we test several imbalance-aware loss functions tailored for skewed distributions, including weighted cross-entropy (wCE), Lovász loss, and focal loss. Details of the evaluation protocol and loss selection procedure are described in Section~\ref{sec:exp_setup}.

In this work, our goal is systematically assess whether Prithvi's pretraining imparts three properties critical for scalable landslide mapping and real-world disaster management: (i) feature reuse, by capturing geospatially rich information such as morphology, vegetation, and soil moisture; (ii) data efficiency, by sustaining strong performance with few labeled samples; and (iii) scalability and generalization, by transferring across diverse geographic regions and sensor configurations. Together, these evaluations establish a comprehensive benchmark of Prithvi-EO-2.0's potential for landslide mapping, clarifying both its current strengths and its remaining limitations.

\begin{figure}[ht]
    \centering
    \includegraphics[width=\linewidth]{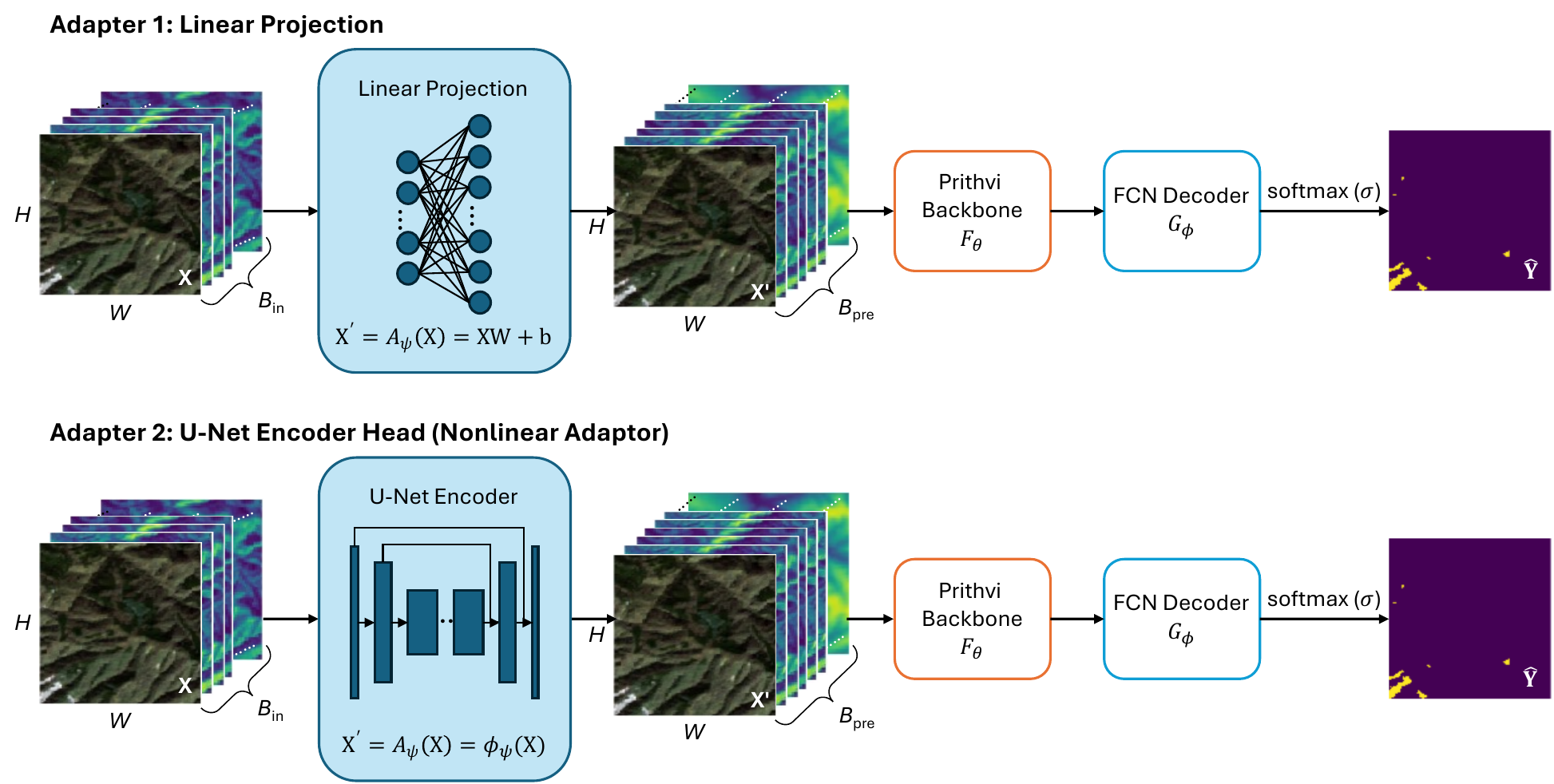}
    \caption{Adapter strategies for band alignment prior to the pretrained Prithvi encoder. Adapter 1: Linear Projection applies a per-pixel affine mapping to project from $B_{\mathrm{in}}$ to the six-band interface $B_{\mathrm{pre}}{=}6$. Adapter 2: U-Net Encoder Head uses a shallow convolutional encoder to capture local spatial context. The projected input $\mathbf{X}'\!\in\!\mathbb{R}^{H\times W\times 6}$ is passed to the Prithvi backbone $F_{\theta}$ (initialized from pretraining) and an lightweight FCN decoder $G_{\phi}$; softmax $\sigma$ applied to logits $\mathbf{S}$ yields the prediction $\hat{\mathbf{Y}}{=}\sigma(\mathbf{S})$. Inputs follow the spectral configurations in Table~\ref{tab:band_configs}.}
    \label{fig:adaptation_strategy}
\end{figure}

\subsection{Band Adaptability}  \label{sec:spectral_band_combinations}

In real-world applications, Earth observation workflows may not align with the six-band configuration used to pretrain Prithvi-EO-2.0 on HLS (B2, B3, B4, B8A, B11, B12). Operational data may instead provide fewer channels, such as RGB-only inputs, or richer ones, such as hyperspectral imagery or multispectral data combined with DEM products. To address this gap, we evaluate \emph{band adaptability}, defined as the ability of a pretrained encoder to maintain segmentation performance when the spectral layout of the input differs from that used during pretraining. Band adaptability directly instantiates the scalability and generalization objective along the sensor axis: aligning one model to heterogeneous band availability across regions for rapid reuse without full retraining. This concept, also referred to as input adaptation \citep{hsu2025geospatial, thoreau2025parameter} or band/channel adaptation \citep{li2025fleximo,li2025hyperfree}, has become a central focus in recent work on geospatial foundation models. 

Prior studies consistently highlight three principles for achieving such adaptability: (i) lightweight mapping layers are often sufficient to reconcile spectral differences, (ii) parameter-efficient alignment modules yield most of the benefit while keeping the backbone stable, and (iii) explicitly or implicitly conditioning the mapping on band identity or wavelength helps preserve physical meaning \citep{thoreau2025parameter,li2025fleximo,li2025hyperfree}. We embed each of these principles into our experimental design. The first is operationalized by contrasting a linear projection, which applies a per-pixel affine mapping into the six-band space, with a shallow U-Net adapter that introduces nonlinear interactions and local spatial context. The second is tested by pairing each adapter with either a frozen or a trainable Prithvi encoder, allowing us to measure whether alignment modules can compensate without disrupting pretrained features. The third is addressed through a spectrum of band configurations that deliberately vary both availability and semantics, ranging from full multisensor fusion to reduced HLS-like subsets, permuted orders, mutual-information-based selections, and minimal RGB plus NIR inputs. Together, these three axes yield a comprehensive grid of conditions that directly translate theoretical principles into controlled experimental factors.

Formally, let $\mathbf{X}\in\mathbb{R}^{H\times W\times B_{\mathrm{in}}}$ denote a multi-band EO patch, where $B_{\mathrm{in}}$ may differ from the six bands used in pretraining ($B_{\mathrm{pre}}=6$). We introduce a lightweight alignment module
\[
A_{\psi}:\ \mathbb{R}^{H\times W\times B_{\mathrm{in}}}\ \rightarrow\ \mathbb{R}^{H\times W\times B_{\mathrm{pre}}},
\]
which projects the input into the six-band interface expected by Prithvi. The projected input is then processed by the pretrained encoder $F_{\theta}$ and segmentation head $G_{\phi}$ defined in the previous section:
\[
\mathbf{Z}=F_{\theta}\big(A_{\psi}(\mathbf{X})\big), \qquad \mathbf{S}=G_{\phi}(\mathbf{Z}), \qquad \hat{\mathbf{Y}}=\sigma(\mathbf{S}).
\]

When the input already matches the six-band pretraining interface ($B_{\mathrm{in}}=B_{\mathrm{pre}}=6$), we bypass $A_{\psi}$ and feed $\mathbf{X}$ directly into the encoder $F_{\theta}$. Adaptation modules are only introduced when the input diverges from this interface in channel numbers.

We evaluate two adapter strategies of $A_{\psi}$ (Figure~\ref{fig:adaptation_strategy}):
\begin{itemize}
    \item \textbf{Linear projection.} Implemented as a per-pixel affine transformation or $1{\times}1$ convolution,
    \[
        A_{\psi}(\mathbf{X})=\mathbf{X}\mathbf{W}+\mathbf{b}, \quad 
        \mathbf{W}\in\mathbb{R}^{B_{\mathrm{in}}\times B_{\mathrm{pre}}},\ \mathbf{b}\in\mathbb{R}^{B_{\mathrm{pre}}},
    \]
    i.e., for each spatial location $(i,j)$ with vector $\mathbf{x}_{ij}\!\in\!\mathbb{R}^{B_{\mathrm{in}}}$, $\mathbf{x}'_{ij}=\mathbf{x}_{ij}\mathbf{W}+\mathbf{b}$. This linearly projects arbitrary spectral inputs into the pretrained six-band space with minimal parameters and no spatial coupling.

    \item \textbf{U-Net encoder head (Nonlinear adapter).} Denoted $\phi_{\psi}$, this shallow convolutional encoder leverages local spatial context and nonlinear interactions between bands and outputs 6 channels:
    \[
	A_{\psi}(\mathbf{X})=\phi_{\psi}(\mathbf{X}) \in \mathbb{R}^{H\times W\times 6}. 
	\]
\end{itemize}

\begin{table}[ht]
	\begingroup
	\renewcommand{\arraystretch}{1.25}
	\caption{Spectral configurations evaluated for band adaptability experiments. Band identifiers follow Sentinel-2 conventions (e.g., B1 = Coastal, B2 = Blue, B3 = Green, B4 = Red, B5 = Red-edge 1, B6 = Red-edge 2, B7 = Red-edge 3, B8 = NIR broad, B8A = NIR narrow, B9 = Water Vapor, B10 = Cirrus, B11 = SWIR1, B12 = SWIR2). Landslide4Sense provides B8 instead of B8A.}
	\centering
	\begin{tabularx}{\textwidth}{>{\raggedright}p{2.8cm} >{\RaggedRight}p{5cm} >{\RaggedRight}X}
	  \toprule
	  \textbf{Configuration} & \textbf{Bands} & \textbf{Rationale} \\
	  \midrule
	  Full (14B) & All Landslide4Sense channels (Sentinel-2 subset + DEM, slope) & Maximal fusion; upper performance bound \\
	  Nine-band (9B) & B2, B3, B4, B8, B11, B12 + B5, B6, B7 & HLS 6-band set plus red-edge bands; commonly used in landslide mapping (spectral only) \\
	  HLS baseline (6B) & B2, B3, B4, B8, B11, B12 & Matches pretraining interface (with B8 in place of B8A) \\
	  HLS shuffled (6B) & Same six bands, \textit{permuted} order & Tests sensitivity to spectral ordering \\
	  MI-6a (6B) & B2, B3, B5, B7, B8, B9 & Selected via mutual information with labels \\
	  MI-6b (6B) & B1, B2, B3, B4, B9, DEM & Alternative mutual-information set including DEM \\
	  RGB+NIR (4B) & B2, B3, B4, B8 & Standard multispectral combination widely used in semantic segmentation \\
	  \bottomrule
	\end{tabularx}
	\label{tab:band_configs}
	\endgroup
  \end{table}
  
To probe adaptability across input settings and the importance of spectral semantics, we design seven spectral configurations that span the range from maximal multisensor fusion to compact 4-band inputs (Table~\ref{tab:band_configs}). These include: (i) a full 14-band setting; (ii) a nine-band configuration combining the six HLS bands with three red-edge bands; (iii) four six-band variants, including the HLS baseline, a permuted-order variant, and two alternatives selected with a mutual information (MI) criterion (see §\ref{sec:mi_criterion}); and (iv) a four-band RGB+NIR setting, widely used in generic semantic segmentation. For each configuration, we train models with both adapters and both backbone settings, creating a factorial design that systematically tests whether lightweight mappings suffice, whether freezing the backbone preserves most of the benefit, and whether band identity and order critically affect adaptation. Training budgets, loss selection, and evaluation metrics follow Section~\ref{sec:exp_setup}.

\paragraph{Mutual information criterion.}\label{sec:mi_criterion}
To construct alternative six-band inputs, we apply a mutual information (MI) criterion to identify the channels most informative about the landslide label. Mutual information is a non-negative measure of dependency between two random variables; it equals zero if and only if the variables are independent, and higher values indicate stronger dependency. Formally, the MI between a spectral band $X_b$ and the binary mask $Y$ is 
\[
I(X_b;Y)=\sum_{x,y}p(x,y)\log\frac{p(x,y)}{p(x)p(y)},
\]
where $x$ denotes a possible pixel intensity (or reflectance value) in band $b$, and $y$ represents the corresponding class label (landslide or non-landslide). The joint probability $p(x,y)$ thus measures how often a given pixel value $x$ and class label $y$ co-occur across the dataset. This formulation captures both linear and nonlinear dependencies between image intensity and class membership.

In practice, we estimated MI using \texttt{mutual\_info\_classif} from the \texttt{scikit-learn} package, a nonparametric $k$-nearest neighbors estimator designed for continuous features and discrete targets. For each $128\times128$ training image, we randomly sampled 4,000 pixel-label pairs (a pixel value from each band together with the corresponding binary mask value). Aggregating across all training images yielded a feature matrix $X \in \mathbb{R}^{N \times 14}$, where $N = 4000 \times$ (number of training images), and a corresponding label vector $Y \in {0,1}^N$. We then computed MI scores between each of the 14 channels and $Y$, producing 14 scores in total. Channels were ranked in descending order of MI, and the top six were chosen as the most informative subset. To assess robustness, we repeated this procedure twice, yielding two distinct six-band configurations (MI-6a and MI-6b in Table~\ref{tab:band_configs}).

\subsection{Data Efficiency and Few-Shot Capability} \label{sec:data_efficiency}

The second property we investigated is \emph{data efficiency}, the ability of a pretrained encoder to sustain strong performance under severe label scarcity. In rapid disaster response, labeled landslide data are often scarce, costly, and slow to produce; inventories are typically reactive and class-imbalanced, and models often need to be adapted to new regions on short notice. A model that maintains accuracy with few labeled samples enables faster, more equitable mapping and timely situational awareness. This directly instantiates the data-efficiency objective outlined in Section~\ref{sec:prithvi_model}, which complements channel adaptability by focusing on the label axis rather than the sensor axis.

Prior work in computer vision and geospatial AI has shown that large-scale pretraining substantially reduces sample complexity by learning semantically rich features from massive unlabeled datasets, thereby providing stronger inductive biases and more stable optimization under label scarcity \citep{he2022masked, oquab2023dinov2, kornblith2019better}. In geospatial contexts, pretraining on multi-temporal, multi-spectral Earth observation data has been shown to capture temporal invariances, spectral correlations, and contextual dependencies that enhance cross-region transferability \citep{cong2022satmae, szwarcman2024prithvi}. Moreover, models that leverage diverse modalities, such as vegetation loss in NIR, soil moisture in SWIR, or terrain attributes from DEMs, have been reported to stabilize decision boundaries and reduce false positives in few-shot scenarios \citep{jakubik2025terramind, xiao2025foundation}. 

Based on these insights, we hypothesize that Prithvi-EO-2.0, pretrained on 4.2M Harmonized Landsat-Sentinel-2 (HLS) image sequences, will retain a higher fraction of its full-data performance compared to task-specific CNNs (U-Net, U-Net++) when trained with limited labels.

To evaluate this hypothesis, we performed experiments on label fractions $k \in \{100, 10, 2.5, 1.25\}\%$. Let $D$ denote the full training set and $D_k \subseteq D$ a single stratified subset containing $k\%$ of training samples (each sample retains its full pixel mask). A fixed random seed was used to construct $D_k$, and the same $D_k$ was used for all models to ensure comparability. We fixed the input to the HLS-baseline six bands (B2, B3, B4, B8, B11, B12; using B8 in place of B8A). For each model $M \in \{\text{Prithvi-EO-2.0-300M}, \text{Prithvi-EO-2.0-600M}, \text{U-Net}, \text{U-Net++}\}$ and each $k$, we fine-tuned $M$ on $D_k$ with identical hyperparameters and recorded the single-run test score on the full testing set.
\[
\mathcal{P}_M(k)\,,
\]
on a higher-is-better segmentation metric (e.g., mIoU). 

We reporteds two complementary, scale-free measures computed relative to each model’s own $100\%$ baseline:
\begin{itemize}
  \item \textbf{Relative Performance Drop (RPD).}
  \[
  \mathrm{RPD}_M(k)=1-\frac{\mathcal{P}_M(k)}{\mathcal{P}_M(100)}.
  \]
  Lower is better; for instance, $\mathrm{RPD}(10)=0.08$ indicates that the model retains $92\%$ of its full-data score with only $10\%$ labels.
  \item \textbf{Aggregated Data Efficiency (DE).}
  \[
  \mathrm{DE}_M=\frac{1}{|K'|}\sum_{k\in K'} \frac{\mathcal{P}_M(k)}{\mathcal{P}_M(100)},\qquad
  K'=\{10,\,2.5,\,1.25\}.
  \]
  Interpreted as average retention under scarce-label regimes, values closer to $1$ indicate stronger few-shot readiness.
\end{itemize}

Altogether, this framework provides a consistent and comparable assessment of data efficiency and few-shot capability, enabling us to evaluate whether geospatial foundation models like Prithvi-EO-2.0 mitigate the ``data-hungry'' limitations of task-specific deep learning models.

\begin{figure}[ht]
	\centering
    \includegraphics[width=\linewidth]{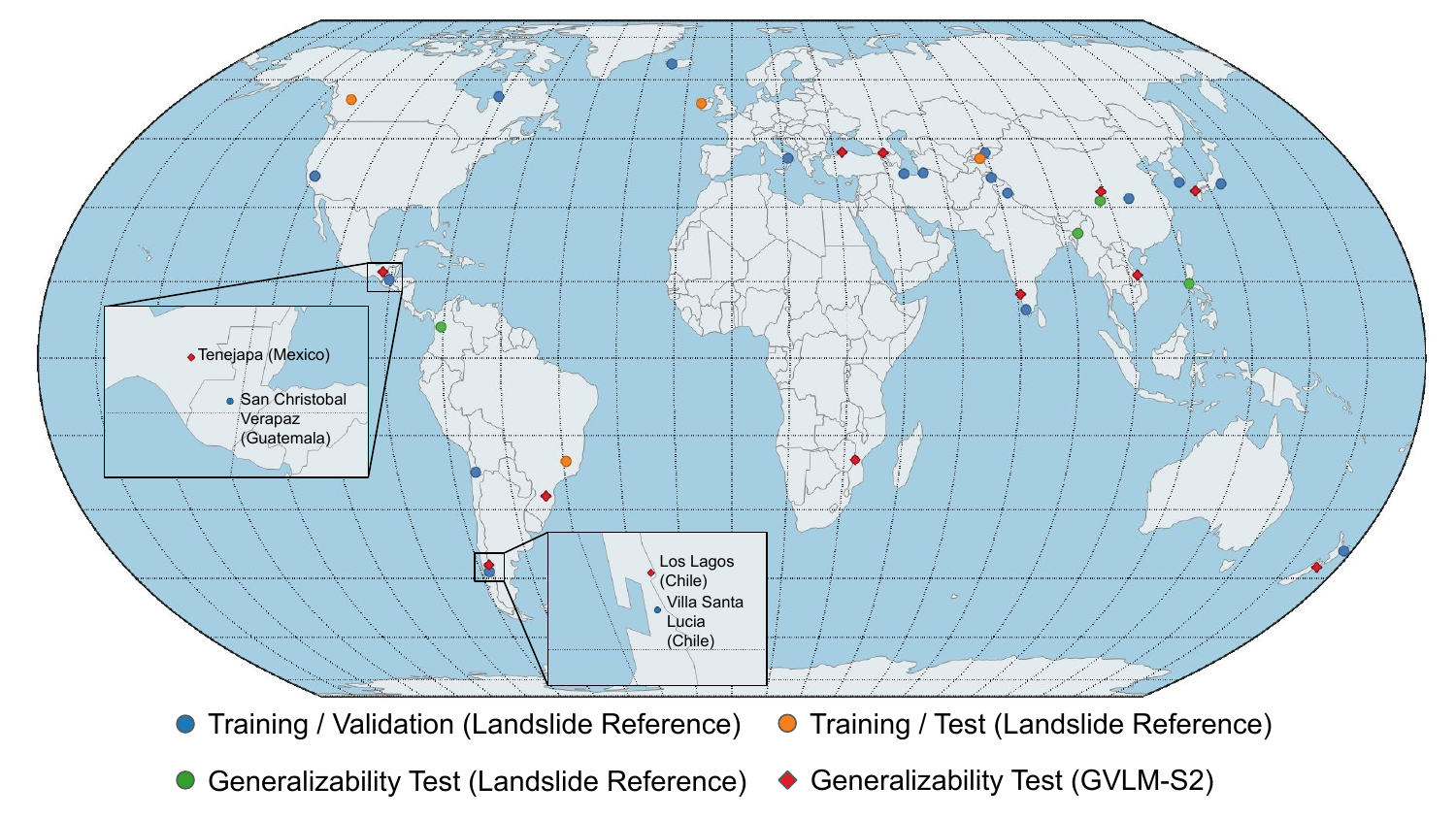}
    \caption{Sites used in the cross-dataset generalization study. Blue/orange: Landslide Reference train/val and test regions; green: Landslide Reference generalizability sites; red diamonds: GVLM-S2 external sites.}
    \label{fig:generalizability_map}
\end{figure}

\begin{figure}[ht]
	\centering
	\includegraphics[width=\linewidth]{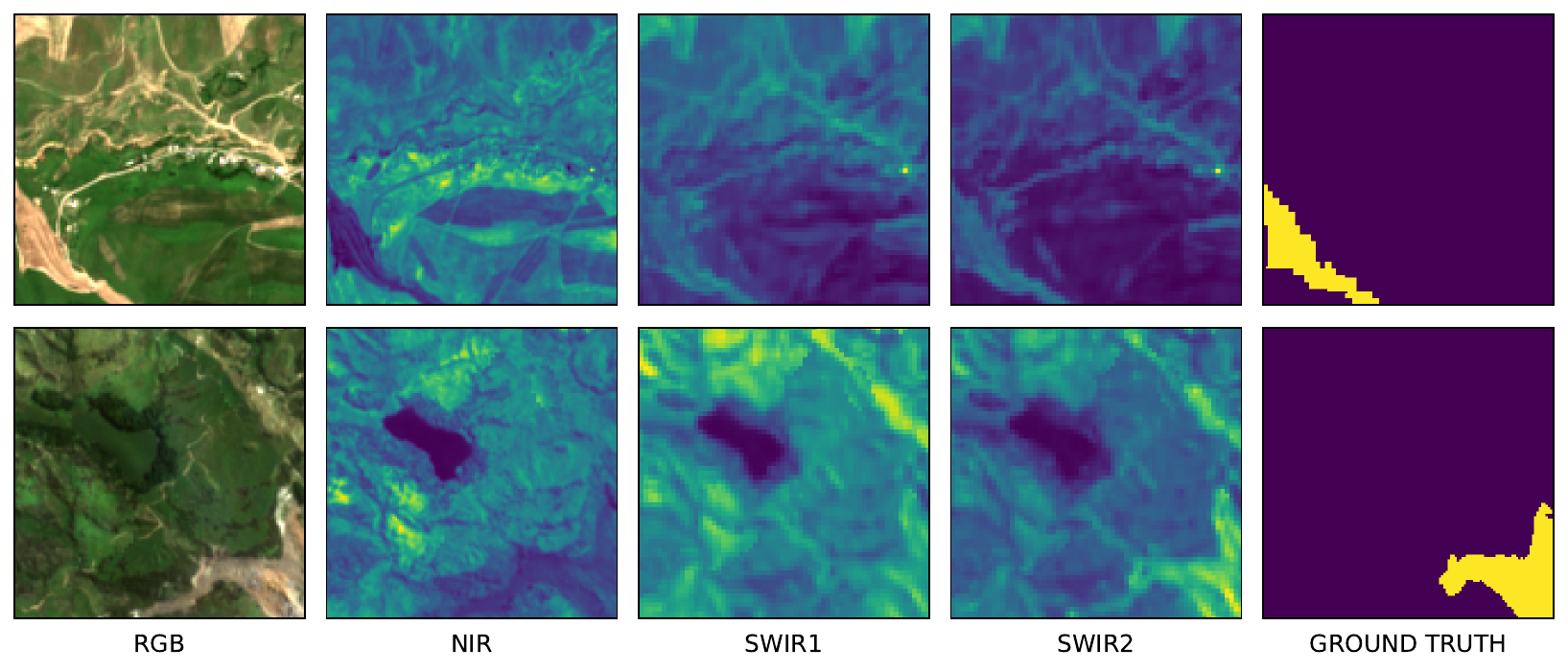} 
	\caption{Sample visualizations from the Landslide Reference dataset (optical-only subset used in this study).}
	\label{fig:lref_samples}
\end{figure}
  
\begin{figure}[ht]
	\centering
	\includegraphics[width=\linewidth]{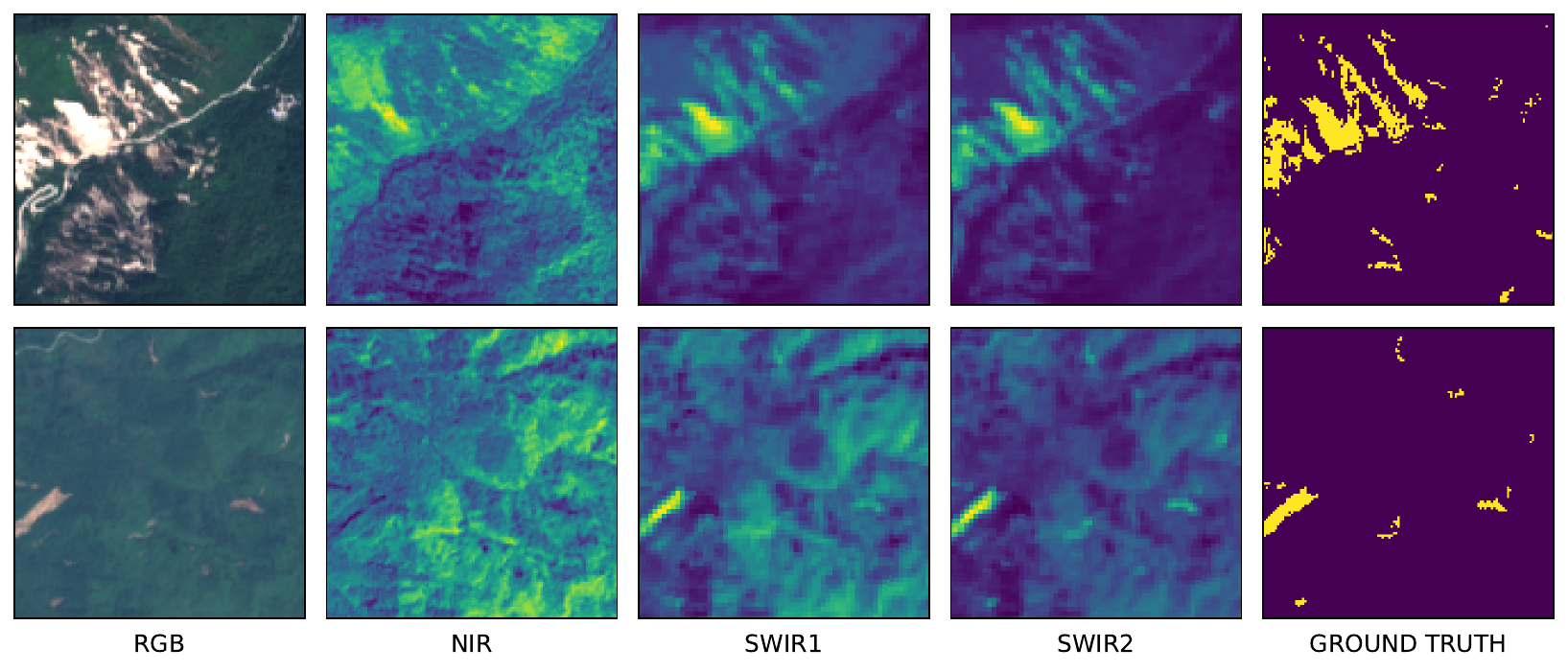} 
	\caption{Sample visualizations from the GVLM-S2 external test set derived from GVLM metadata.}
	\label{fig:gvlm_s2_samples}
\end{figure}
 
\subsection{Model Generalizability Across Datasets and Geographies} \label{sec:generalizability}

The third property we examined was \emph{generalizability}: whether a model fine-tuned on one landslide dataset transferred to geographically and institutionally distinct data without any additional tuning. In contrast to band adaptability (sensor axis) and data efficiency (label axis), this experiment focused on the domain axis. The Landslide4Sense dataset was not used because it lacked site coordinates, which prevented us from constructing geographically disjoint targets. Instead, we relied on two datasets with explicit location information for each sample: the Landslide Reference dataset \citep{orynbaikyzy2025landslide} and an extended Global Very-High-Resolution Landslide Mapping (GVLM)dataset \citep{zhang2023cross}. The geographical distributions of these datasets are shown in Figure~\ref{fig:generalizability_map}.

We trained the GeoFM Prithvi-EO on the Landslide Reference dataset \citep{orynbaikyzy2025landslide}, which provides $128{\times}128$ pixel samples with predefined splits of 355 training, 88 validation, and 76 test samples, along with a generalizability split of 44 samples from unseen regions. Although the dataset includes Sentinel-1 SAR (pre/post) and Sentinel-2 optical (pre/post) data, we used only Sentinel-2 to align with Prithvi-EO-2.0's optical pretraining and to avoid confounding modality shifts. Sample visualizations are provided in Figure~\ref{fig:lref_samples}.

To broaden out-of-domain coverage beyond the generalizability test dataset provided by the same authors \citep{orynbaikyzy2025landslide}, we further evaluated on GVLM-S2, an external Sentinel-2 dataset we independently derived from the GVLM metadata \citep{zhang2023cross}. Using the landslide polygons labeled within GVLM, we downloaded Sentinel-2 L1C imagery corresponding to the timestamps of each landslide event for 11 regions, explicitly excluding any region that appeared in the Landslide Reference training dataset. We then resampled the provided labels to the Sentinel-2 grid, yielding 56 external samples. Examples are shown in Figure~\ref{fig:gvlm_s2_samples}.

Our hypothesis is that geospatial foundation model Prithvi-EO-2.0 yields a stronger generalizability capability than task-specific domains due to the pre-training based on large amount of geospatial data. To verify this hypothesis, we fine-tuned each model $M \in \{\text{Prithvi-EO-2.0-300M}, \text{Prithvi-EO-2.0-600M}, \text{U-Net}, \text{U-Net++}\}$ on the Landslide Reference training dataset and selected checkpoints by validation loss on its validation dataset. Inputs were fixed to the HLS-baseline six bands (B2,B3,B4,B8,B11,B12) with B8 used in place of B8A. Per-band standardization statistics were computed on the training split only and reused at inference for all targets to avoid data leakage. 

The chosen checkpoint was then evaluated \emph{as is} on three test datasets: 

\begin{itemize}

\item an in-domain test set $\mathcal{T}{\mathrm{in}}$ (the Landslide Reference test set), where in-domain means the testing dataset came from the same geographical regions as the Landslide Reference training data (refer to Figure~\ref{fig:gvlm_s2_samples} for the distribution of training and testing datasets); 
\item an out-of-domain set $\mathcal{T}{\mathrm{gen}}$ (the Landslide Reference generalizability set from unseen regions); and 

\item an external out-of-domain set $\mathcal{T}_{\mathrm{ext}}$ (GVLM-S2), which did not geographically overlap with the Landslide Reference training dataset. 
\end{itemize}

Primary metrics followed Section~\ref{sec:exp_setup} (e.g., mIoU; higher is better). In addition to absolute scores, we summarized transfer performance with retention ratios.
\[
\mathrm{R}_M(\mathcal{T})=\frac{\mathcal{P}_M(\mathcal{T})}{\mathcal{P}_M(\mathcal{T}_{\mathrm{in}})},
\qquad \mathcal{T}\in\{\mathcal{T}_{\mathrm{gen}},\,\mathcal{T}_{\mathrm{ext}}\},
\]
which tells what fraction of the in-domain performance is preserved on an out-of-domain (OOD) target. 

This design isolates cross-dataset generalization, where models were tuned once on Landslide Reference (optical-only) and then assessed on geographically distinct tiles from the same dataset as well as on an external set curated from independent metadata. Together with the previous experiments, this completed our three-axis evaluation across sensor, label, and domain.

\section{Experiments \& Results}

\subsection{Experimental Setup and Evaluation Metrics} \label{sec:exp_setup}

All models were implemented in PyTorch \citep{paszke2019pytorch} and trained on NVIDIA RTXA5000 GPUs (24,GB each), using two GPUs per run. We followed non-overlapping train/validation/test splits as described in Section \ref{sec:dataset} for Landslide4Sense, and the dataset-specific splits for the cross-dataset generalization experiments on Landslide Reference and GVLM-S2 (see Section~\ref{sec:generalizability}). Unless otherwise noted, we selected the checkpoint that attained the lowest validation loss matching the training loss, which prioritized generalization over overfitting.

Input sizes were $128{\times}128$ pixels. Each channel was standardized using the mean and standard deviation computed on the training split only, and the same statistics were reused at inference for validation/test and external targets to avoid leakage. We applied online augmentation with random horizontal and vertical flips; no test-time augmentation was used. When the input already matched Prithvi's six-band pretraining interface ($B_{\mathrm{in}}=6$), we bypassed the alignment module and fed the input data directly to the encoder. Alignment was introduced only when $B_{\mathrm{in}}\neq 6$, as detailed in Section~\ref{sec:spectral_band_combinations}.

Optimization was kept uniform across models and losses for fair comparison: we fine-tuned for 100 epochs with batch size 8 using Adam (learning rate $1.0\times10^{-5}$, weight decay $=0$) without early stopping. 

To study interactions with extreme class imbalance and region-level overlap, we trained each model independently with three losses:
\begin{itemize}
	\item \textbf{Weighted cross-entropy (wCE).} We used fixed, a priori class weights $(w_{\text{bg}}, w_{\text{ls}})=(2,8)$ (background:landslide $=$ 2:8) to place greater emphasis (e.g., 80\%) on the correct detection of landslide pixels relative to background. This also meant the model will receive a higher penalty for underdetecting landslide pixels. For logits $\mathbf{z}$ and label $y\in\{\text{bg},\text{ls}\}$, $\mathcal{L}_{\mathrm{wCE}}=-\,w_y\log\!\big(\mathrm{softmax}(\mathbf{z})_y\big)$.
    \item \textbf{Lovász-Softmax} \citep{berman2018lovasz}. Unlike cross-entropy, which measures pixel-wise correctness, Lovász-Softmax directly optimizes the Jaccard index (IoU), a region-based overlap metric commonly used for segmentation. By aligning the loss with IoU, it encourages predictions with better overall shape and boundary quality. While not explicitly weighting classes, it indirectly mitigates imbalance by giving stronger influence to small positive regions in IoU optimization. 
    \item \textbf{Focal loss} \citep{lin2017focal}. It modifies cross-entropy by multiplying each term with a factor $(1-p_t)^\gamma$, where $p_t$ is the model's predicted probability for the true class. When $p_t$ is large (easy, well-classified pixel), the factor becomes small, so its loss contribution is down-weighted. When $p_t$ is small (hard, misclassified pixel), the factor stays close to 1, so the loss remains high. The prevents the abundant background pixels from dominating training and shifts learning toward rare and difficult landslide pixels. 
\end{itemize}

Each model-loss pair used the same initialization and schedule, and model selection on the validation set used the corresponding loss to avoid selection bias. Inference was tile-by-tile with a fixed probability threshold of $0.5$ on the landslide class.

We reported complementary, thresholded metrics aligned with operational mapping, computed from the global confusion matrix aggregated over all test tiles:
\begin{itemize}
    \item \textbf{Mean Intersection over Union (mIoU):} the mean of per-class IoUs (landslide and background); primary quality indicator for segmentation under imbalance.
    \item \textbf{F1 score:} harmonic mean of precision and recall; highlights the balance between missed detection (false negative) and false alarms (false positive).
    \item \textbf{Precision / Recall:} fraction of predicted landslide pixels that are correct / fraction of true landslide pixels detected.
    \item \textbf{Mean accuracy (mAcc):} Overall pixel accuracy across classes; intuitive but background-dominated, so we treat it as complementary to mIoU/F1.
\end{itemize}

\subsection{Baseline Performance Evaluation} \label{sec:baselineeval}

Our first experiment reported the benchmark performance of the Prithvi-EO-2.0 model in comparison to other deep learning models. The experiments were conducted on the full Landslide4Sense dataset using the same train/validation/test split guidelines to ensure consistency. We compared four families of landslide segmentation models: (i) task-specific CNNs (U-Net, U-Net++), (ii) general-purpose Vision Transformers (SegFormer, SwinV2-B), (iii) Segment Anything Model (SAM)-adapted landslide models (TransLandSeg), and (iv) other geospatial foundation models (GeoFMs: SatMAE, TerraMind-base). This design enabled like-for-like comparisons across architectural families (CNN vs. Transformer), task specialization (task-specific vs. foundation models), and pretraining paradigms (generic vs. EO-native). 

Task-specific CNNs are classic, fully supervised encoders for EO segmentation that learn features from scratch (or from generic image pretraining) and emphasize locality. We included U-Net and U-Net++ (ResNet50 backbones) as long-standing baselines widely used in landslide and other mapping tasks. The second class is general-purpose Vision Transformers, which represent mainstream cutting-edge models for image analysis and also form the basis of many geospatial foundation models. Vision Transformers achieve state-of-the-art performance because of their ability to capture long-range context and dependencies through self-attention, making detection less constrained by fixed local windows that CNN use. From this class, we trained SegFormer and SwinV2-B and reported their performance. 

The third class is SAM-adapted models, which transfer large vision foundation models trained on natural RGB imagery (e.g., SAM) to landslide segmentation. SAM (Segment Anything Model) is a powerful, general-purpose segmentation backbone that can be adapted to diverse downstream tasks. Since it was designed specifically for segmentation, it provides strong performance. In particular, we compared TransLandSeg, which is built upon SAM and tailored for landslide detection. Finally, we compared Prithvi-EO-2.0 (300M and 600M) with other geospatial foundation models, including SatMAE and TerraMind-base. These models were pretrained on large EO datasets with multispectral (and often multitemporal) inputs to encode geospatial priors for improved performance in geospatial analysis. TerraMind-base and Prithvi-EO-2.0 (300M/600M) directly target feature reuse. If pretraining effectively captures cues related to morphology, vegetation, and terrain conditions, these models should yield higher mIoU/F1 under identical fine-tuning settings.

All models share identical splits, preprocessing and augmentations, optimization settings, and checkpoint selection by lowest validation loss (Section~\ref{sec:exp_setup}). To minimize modality confounders, inputs were standardized to an HLS-like six-band optical stack on Landslide4Sense (B2, B3, B4, B8, B11, B12, with B8 replacing HLS B8A). Architectures with mismatched input interfaces were adapted accordingly, except TransLandSeg, which remains RGB-only.

\begin{table}[ht]
	\begin{threeparttable}
	\caption{Baseline performance on the Landslide4Sense test set (100\% training data). Unless noted, inputs are six optical bands (B2, B3, B4, B8, B11, B12); the best validation-loss checkpoint per model/loss is reported. Metrics are computed on the global confusion matrix (higher is better).} \label{tab:baseline}
	\centering
	\begin{tabularx}{\textwidth}{>{\RaggedRight}p{3.7cm} >{\centering\arraybackslash}p{1.7cm} *{5}{>{\centering\arraybackslash}X}}
		\toprule
		\textbf{Model} & \textbf{Loss Function} & \textbf{mIoU} & \textbf{F1 Score} & \textbf{Precision} & \textbf{Recall} & \textbf{mAcc} \\
		\midrule
		U-Net & wCE & 70.44 & 59.66 & 56.17 & 63.63 & 98.37 \\
        U-Net++ & wCE & 68.96 & 57.89 & 50.19 & 68.38 & 98.03 \\ 
        SegFormer & wCE & 69.89 & 59.12 & 49.51 & 73.37 & -- \\
        SwinV2-B & wCE & 67.77 & 58.28 & 47.09 & \textbf{76.43} & 97.88 \\ 
		TransLandSeg  & BCE + Dice & 68.30 & 55.39 & 55.23 & 55.55 & 98.31 \\ 
        SatMAE (Multi-Spectral) & wCE & 67.48 & 53.72 & 54.00 & 53.44 & 98.26 \\ 
        TerraMind (base) & focal & 69.77 & 58.12 & \textbf{65.95} & 51.95 & \textbf{98.59} \\ 
		\midrule
        Prithvi-EO-2.0-300M & Lovász & \textbf{71.30} & \textbf{60.70} & 57.14 & 64.73 & 98.42 \\
        Prithvi-EO-2.0-600M & wCE & 70.41 & 58.56 & 51.07 & 68.61 & 98.17 \\ 
		\bottomrule
	\end{tabularx}
	\end{threeparttable}
\end{table}

Table \ref{tab:baseline} illustrates the results. Note that the result for TransLandSeg was reproduced by retraining with the shared model code using the standard train/validation/test split of Landslide4Sense. The result reported in the TransLandSeg paper, however, was based on a different split method (a 70/30 train/test split on the Landslide4Sense training set). Inputs were limited to RGB only due to constraints of the SAM backbone. 

Across model families, we observed distinct precision–recall profiles rather than large gaps in mIoU. Task-specific CNNs performed solidly on mIoU. U-Net achieved 70.44 mIoU and 59.66 F1 (56.17 Precision / 63.63 Recall), and U-Net++ achieved 68.96 mIoU and 57.89 F1 (50.19 Precision / 68.38 Recall). In both cases, recall exceeded precision, indicating over-segmentation (more false positives) despite reasonable pixel-wise overlap. This result highlights the challenges of distinguishing landslide scars from spectrally similar terrain.

General-purpose Vision Transformers showed an even stronger recall-heavy trade-off. SegFormer achieved 69.89 mIoU and 59.12 F1 with 49.51 precision versus 73.37 recall, while SwinV2-B attained the highest recall in the table (76.43) but the lowest precision (47.09), yielding 67.77 mIoU and 58.28 F1. This suggests that global attention helps recover elongated, fragmented landslide scars that exceed the typical processing window of a CNN. However, without EO-native pretraining, these models struggle to suppress spurious detections. The SAM-adapted TransLandSeg (RGB-only, per the SAM backbone) performed lower overall, with 68.30 mIoU and 55.39 F1 and roughly balanced 55.23 precision / 55.55 recall, highlighting the cost of losing multi-spectral cues.

Among geospatial foundation models, the patterns are clearer. SatMAE (multi-spectral) underperformed with 67.48 mIoU and 53.72 F1 (54.00 precision / 53.44 recall). TerraMind (base) delivered the best precision in the benchmark (65.95) but with low recall (51.95), yielding 69.77 mIoU and 58.12 F1. Prithvi-EO-2.0 set the overall baseline. The 300M variant achieved the highest mIoU and F1 (71.30 / 60.70, with balanced 57.14 precision / 64.73 recall), and the 600M variant remained competitive (70.41 / 58.56; 51.07 precision / 68.61 recall). Since most models here use six optical bands (RGB+NIR+SWIR1+SWIR2) while TransLandSeg is RGB-only, these gains point to genuine feature reuse from EO-native, multi-spectral pretraining, which captures important environmental conditions and spatial context, yielding better precision–recall balance and higher segmentation accuracy. (Note that mAcc is uniformly high across all models and less diagnostic under class imbalance.)

\subsection{Band Adaptability with Varied Spectral Inputs}

This section summarizes the results and findings on Prithvi-EO's ability adapt to heterogeneous band combinations when data sources differ from its pretrained EO inputs (i.e., the six fixed bands of Prithvi), as often occurs in real-world disaster scenarios. Table~\ref{tab:prithvi_channel_adapt} compares the two best-performing models based on results in Section \ref{sec:baselineeval}, U-Net and Prithvi-EO-2.0-300M, across spectral configurations (14B, 9B, 6B, 4B), adapter heads, and encoder tuning strategies. As shown in the table, when inputs were matched, Prithvi-EO-2.0 300M consistently outperformed the task-specific U-Net. Under the 6B setting, U-Net achieved 70.44 mIoU and 59.66 F1, whereas Prithvi-EO-2.0 300M reached 71.30 mIoU and 60.70 F1 with full tuning (tuned on all trainable parameters without backbone frozen), indicating that EO-native pretraining provides reusable geospatial features. At 14B, Prithvi already surpassed U-Net with a frozen encoder (71.23 mIoU and 61.86 F1 vs. 61.20 and 46.42), and end-to-end tuning further increased performance to 73.62 mIoU and 65.42 F1. This shows that adding an adapter to handle all 14 bands was able to generate enhanced 6-band features, boosting overall performance. Using this adapter approach, we observed the best model performance of 73.62 mIoU across all testing scenarios on the full Landslide4Sense dataset.

\begin{table}[ht]
	\caption{Band adaptability results across band configurations, adapter heads, and tuning strategies.
	Segmentation performance is reported for U-Net and Prithvi-EO-2.0-300M under four spectral settings (14B, 9B, 6B, 4B, see Table~\ref{tab:band_configs}). The 6B configuration is HLS-like (B2, B3, B4, B8, B11, B12, with B8 replacing HLS B8A). Prithvi models are paired with either a linear projection or U-Net adapter and evaluated with frozen or full encoder tuning. Frozen means the backbone was kept fixed during fine-tuning, while Full means all parameters were tuned with no modules frozen.}
	\centering
    \begin{tabularx}{\textwidth}{>{\RaggedRight\arraybackslash}p{2.1cm} *{8}{>{\centering\arraybackslash}X}}
		\toprule
		\textbf{Model} & \textbf{Bands} & \textbf{Adapter} & \textbf{Tuning Strategy} & \textbf{mIoU} & \textbf{F1} & \textbf{Precision} & \textbf{Recall} & \textbf{mAcc} \\
		\midrule
        U-Net & 6B & -- & Full & 70.44 & 59.66 & 56.17 & 63.63 & 98.37 \\ 
        U-Net & 14B & -- & Full & 61.20 & 46.42 & 60.68 & 37.59 & 97.96 \\
		\midrule
        Prithvi-EO-2.0 & 14B & U-Net & Frozen & 71.64 & 61.86 & 59.92 & 63.93 & 98.51 \\
        Prithvi-EO-2.0 & 14B & U-Net & Full & \textbf{73.62} & \textbf{65.42} & 63.36 & \textbf{67.62} & \textbf{98.65}  \\ 
        Prithvi-EO-2.0 & 14B & Linear & Frozen & 71.08 & 60.82 & 59.31 & 62.42 & 98.48 \\
        Prithvi-EO-2.0 & 14B & Linear & Full & 70.43 & 59.52 & 61.06 & 58.06 & 98.51 \\
        Prithvi-EO-2.0 & 9B & U-Net & Frozen & 68.77 & 56.27 & 58.01 & 54.63 & 98.40 \\ 
        Prithvi-EO-2.0 & 9B & Linear & Frozen & 63.60 & 46.35 & 35.46 & 66.87 & 97.08 \\
        Prithvi-EO-2.0 & 6B & -- & Frozen & 66.21 & 51.53 & 43.38 & 63.44 & 97.75 \\
        Prithvi-EO-2.0 & 6B & -- & Full & 71.30 & 60.70 & 57.14 & 64.73 & 98.42  \\
        Prithvi-EO-2.0 & 4B & U-Net & Frozen & 65.66 & 50.46 & 41.33 & 64.77 & 97.60\\ 
        Prithvi-EO-2.0 & 4B & Linear & Frozen & 59.65 & 34.68 & \textbf{65.14} & 23.63 & 98.32 \\ 
		\bottomrule
	\end{tabularx}
	\label{tab:prithvi_channel_adapt}
\end{table}

In addition, we observed that adapter choice strongly influenced the precision–recall balance. With 9B inputs and a frozen encoder, a U-Net adapter achieved much higher mIoU/F1 than a linear projection (68.77 / 56.27 vs. 63.60 / 46.35), while the linear mapping drove recall much higher (66.87) at the expense of precision and F1, reflecting issues of over-segmentation. Conversely, with 4B inputs, the linear mapping did the opposite (higher precision, much lower recall; 65.14 vs. 23.62). Both cases reduced F1 relative to the U-Net adapter. These patterns suggest that lightweight, nonlinear alignment provides an effective, parameter-efficient mechanism to reconcile band mismatches, and that frozen encoders can preserve much of the benefit when compute or labels are limited.

\begin{table}[ht]
	\caption{Segmentation performance of Prithvi-EO-2.0-300M across spectral band configurations. All experiments use a frozen encoder with a U-Net adapter, except for 6-band inputs, which matches the pretraining interface and requires no adapter. Band identifiers follow Sentinel-2 conventions (B1-B12; see Table~\ref{tab:band_configs} for details).}
	\centering
    \begin{tabularx}{\textwidth}{>{\RaggedRight\arraybackslash}p{2.7cm} >{\RaggedRight\arraybackslash}p{4.6cm} *{5}{>{\centering\arraybackslash}X}}
		\toprule
		\textbf{Configuration} & \textbf{Bands} & \textbf{mIoU} & \textbf{F1} & \textbf{Precision} & \textbf{Recall} & \textbf{mAcc} \\
		\midrule
		Full (14B) & All Landslide4Sense channels & \textbf{71.64} & \textbf{61.86} & 59.92 &\textbf{ 63.93 }& 98.51 \\
        Nine-band (9B) & HLS baseline + B5, B6, B7 & 68.77 & 56.27 & 58.01 & 54.63 & 98.40 \\ 
        HLS baseline (6B) & B2, B3, B4, B8, B11, B12 & 66.21 & 51.53 & 43.38 & 63.44 & 97.75 \\
        HLS shuffled (6B) & Same six bands, permuted order & 60.62 & 39.17 & 30.45 & 54.89 & 96.79 \\ 
        MI-6a (6B) & B2, B3, B5, B7, B8, B9 & 70.08 & 58.79 & \textbf{62.42} & 55.56 & \textbf{98.53 }\\ 
        MI-6b (6B) & B1, B2, B3, B4, B9, DEM & 70.64 & 59.25 & 60.46 & 58.08 & 98.49 \\
		\bottomrule
	\end{tabularx}
	\label{tab:band-combos}
\end{table}

We conducted further experiments to better understand band semantics by isolating the effect of spectral choice under a fixed model setting: a frozen Prithvi-EO encoder with a U-Net adapter. Different number of bands and band combinations served as the input of the model. Note no adapter was used in the experiment with 6-band input because the input band numbers aligned with input of Prithvi. The experimental results are shown in Table ~\ref{tab:band-combos}. First, the full 14B fusion gives the best overall performance (71.64 mIoU, 61.86 F1). The HLS 6B baseline is lower (66.21 mIoU, 51.53 F1), note in this case, while simply permuting those six bands (HLS shuffled) drops sharply to 60.62 mIoU and 39.17 F1, showing that band identity and ordering, not just channel count, matter. 

The informed six-band sets (MI-6a, MI-6b) were constructed by ranking spectral channels to the landslide label using mutual information (MI, see §\ref{sec:mi_criterion}), a measure of dependency between variables where higher values indicate stronger association. Both MI-6a and MI-6b achieved over 70 mIoU and an F1 score of about 59, nearly closing the gap to the full 14B input and clearly outperformed the HLS 6B baseline (66.21 mIoU / 51.53 F1). Although the MI favors channels that carry the strongest statistical dependency with the landslide label, the resulting subsets also align with physically relevant signals: MI-6a emphasizes NIR and SWIR bands (B5, B7, B8, B9) alongside visible channels (B2, B3), highlighting vegetation and moisture sensitivity, while MI-6b combines optical bands (B1-B4, B9) with topography (DEM), linking directly to slope and hydrological controls. Thus, MI-guided selections yield subsets that are both statistically grounded and physically interpretable, rather than arbitrary combinations. 

Overall, the results support three takeaways. (i) EO-native pretraining gives Prithvi strong band adaptability, clearly outperforming task-specific CNNs under matched inputs. (ii) A lightweight nonlinear adapter is preferred over a linear projection because it avoids pathological precision–recall shifts and yields higher F1; a frozen encoder plus adapter is already strong when computing resources are limited, while full tuning adds headroom. (iii) Band identity and ordering matter, but the optimal band set is task-aligned rather than tied to the pretraining interface. Informed six-band selections (MI-6a/MI-6b), chosen for their statistical dependency with the landslide label, nearly match the 14B result and outperform the HLS 6B baseline, while also aligning with semantically meaningful signals such as vegetation, soil, and moisture. By contrast, simply permuting the HLS bands collapses F1 to 39.17. This shows that Prithvi’s encoder can repurpose features for task-aligned inputs, but performance degrades when band identity or order are disrupted. In addition, band count alone is insufficient. In practice, one should select mission-relevant bands, maintain a consistent channel order, and, when the number or composition of input bands differs from the pretraining configuration, use a nonlinear adapter to map them to the encoder.

\begin{figure}[ht]
	\centering
	\begin{subfigure}{0.6\textwidth}
	\includegraphics[width=\linewidth]{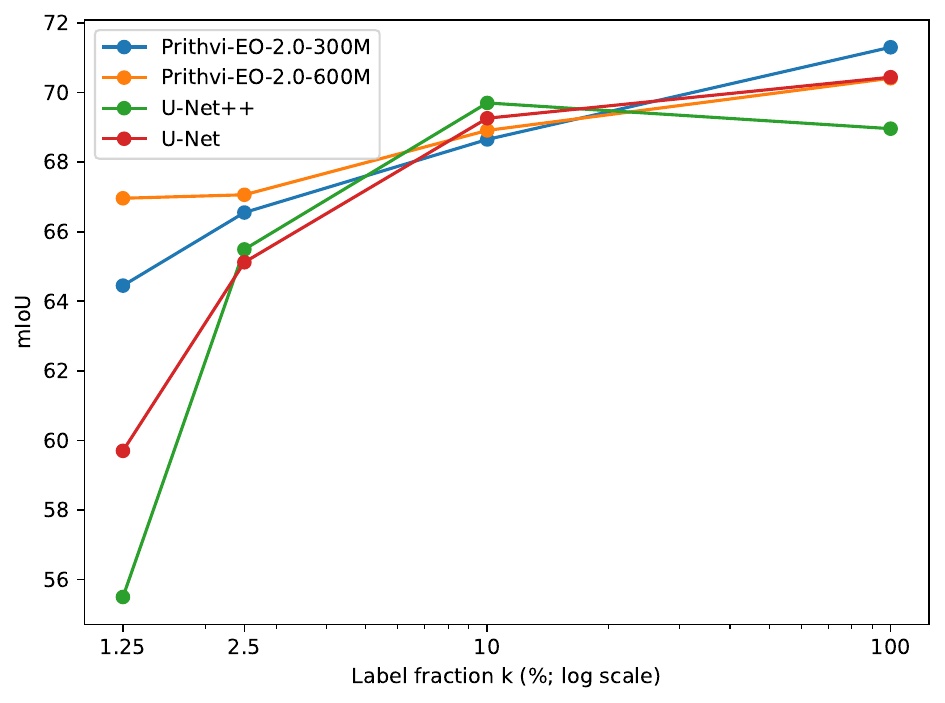}	
	\caption{} \label{fig:exp_data_efficiency_sample_efficiency}
	\end{subfigure}
	\begin{subfigure}{0.49\textwidth}
	\includegraphics[width=\linewidth]{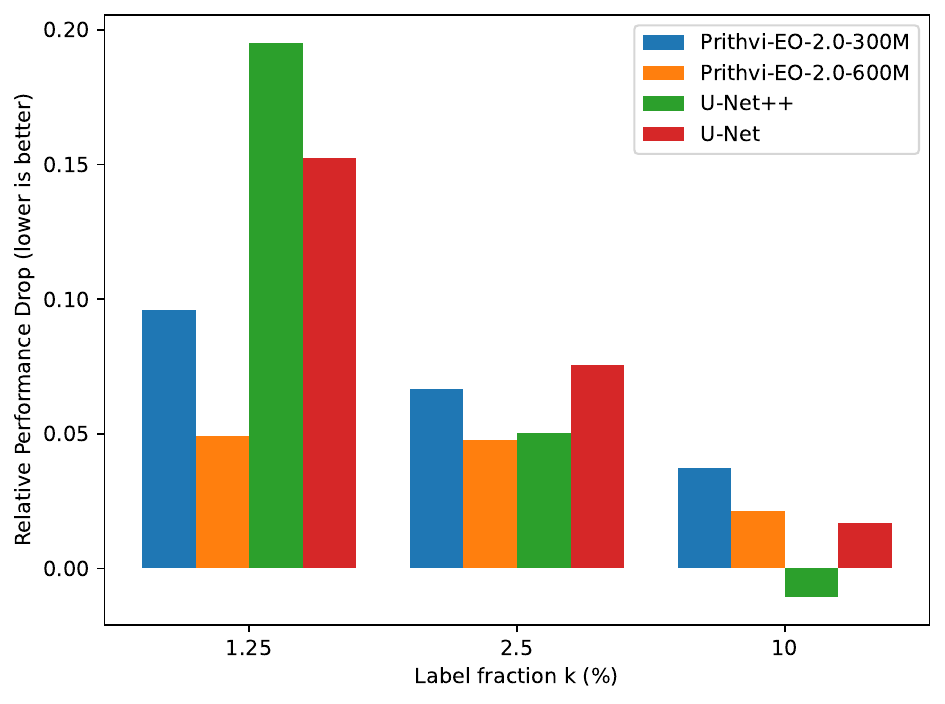}
	\caption{} \label{fig:exp_data_efficiency_rpd_bars}
	\end{subfigure}
	\begin{subfigure}{0.49\textwidth}
	\includegraphics[width=\linewidth]{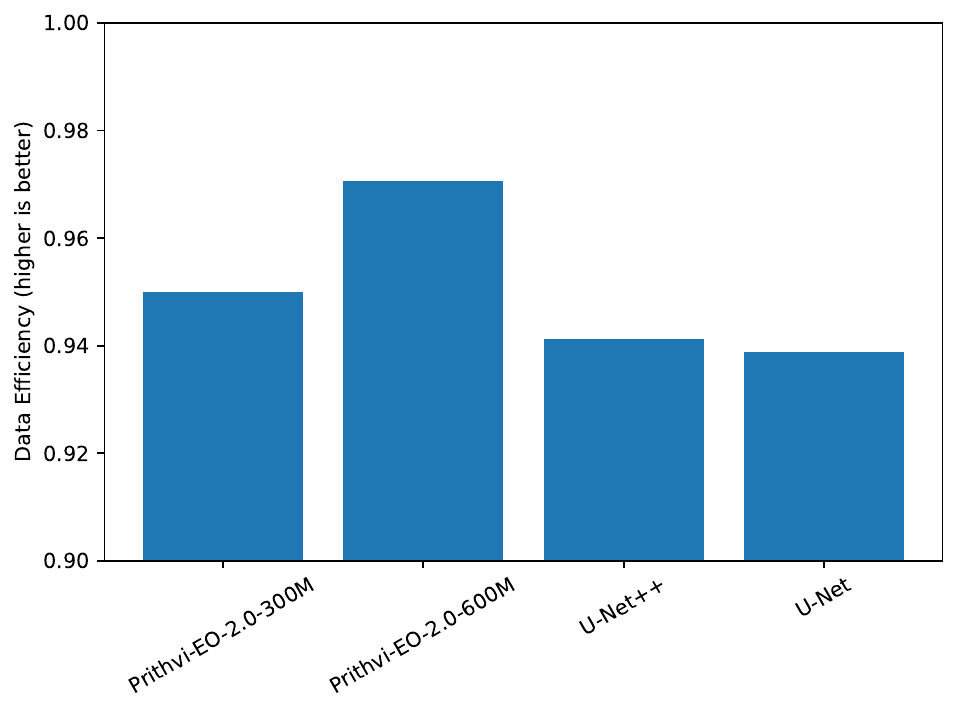}
	\caption{} \label{fig:exp_data_efficiency_de_bars}
	\end{subfigure}
	\caption{ (a) mIoU versus labeled fraction $k\in\{1.25,2.5,10,100\}$ (log-scaled $x$), comparing Prithvi-EO-2.0 (300M, 600M) with U-Net++ and U-Net. (b) Relative performance drop $\mathrm{RPD}_M(k)=1-\mathcal{P}_M(k)/\mathcal{P}_M(100)$ (lower is better). (c) Aggregate data efficiency $\mathrm{DE}_M=\tfrac{1}{3}\sum_{k\in\{10,2.5,1.25\}}\mathcal{P}_M(k)/\mathcal{P}_M(100)$ (higher is better). All models are fine-tuned with identical hyperparameters on the same $D_k$ split (single run per setting); inputs use the six-band HLS interface (B2, B3, B4, B8, B11, B12).}
	\label{fig:exp_data_efficiency}	
\end{figure}

\subsection{Data Efficiency and Few-Shot Capabilities} 

This section further assesses the performance of the Prithvi-EO (300M and 600M) in comparison with another top-performing U-Net class models (UNet and UNet++). The experiments were conducted at a reduced size of training data, from full dataset to a very small percent (at 1.25\% and approximately 50 samples). Figure~\ref{fig:exp_data_efficiency_rpd_bars} demonstrates the results. 

As shown in Figure~\ref{fig:exp_data_efficiency_rpd_bars}(a), across single-seed runs at label fractions $k\in\{1.25,2.5,10,100\}\%$, both Prithvi variants degrade gracefully as labels become scarce, while the task-specific CNNs exhibit sharper drops at the most extreme scarcity. At $1.25\%$ labels, Prithvi-EO-2.0-600M attains 66.96 mIoU, followed by Prithvi-300M at 64.45; the CNN baselines trail (U-Net: 59.70; U-Net++: 55.50). Thus, under the strictest regime the 600M model enjoys margins of 7-12(\%) mIoU over U-Net and U-Net++, indicating stronger few-shot readiness.

At $2.5\%$, the mIoU gap narrows (Prithvi-600M: 67.06; Prithvi-300M: 66.55; U-Net++: 65.49; U-Net: 65.12), suggesting that modest label increases help CNNs close in but do not erase the advantage of EO-native pretraining. By $10\%$ labels, all models cluster near 69-70 mIoU. The slight edge for U-Net++ at $10\%$ coincides with a small negative RPD (-0.0107), i.e., the single run at $10\%$ exceeds its own $100\%$ baseline by \(\approx\)0.7 points. This non-monotonicity is consistent with single-seed variance or minor overfitting when the entire dataset is used for training.

Evaluated relative to each model's full-data score (Figure~\ref{fig:exp_data_efficiency_rpd_bars}), Prithvi-600M remains in the low single digits, showing the most stable retention across label scarcity: $\mathrm{RPD}(1.25)=0.049$, $\mathrm{RPD}(2.5)=0.0476$, and $\mathrm{RPD}(10)=0.0213$, which means the model yields performance retention rates at 95.1\%, 95.2\%, and 97.9\%  when 1.25\%. 2.5\%, and 10\% of the data are used for training respectively. The CNNs exhibit a clear performance drop in a data-scarce situation (i.e., only $1.25\%$ of data is available for training). U-Net++ shows a slightly negative RPD at $10\%$, where its single run marginally exceeds its own $100\%$ baseline, an effect attributable to single-seed variability rather than a systematic advantage.

Figure~\ref{fig:exp_data_efficiency_de_bars} aggregates scarce-label behavior through the data-efficiency score (DE). The ranking is consistent with Figure~\ref{fig:exp_data_efficiency_rpd_bars}: Prithvi-600M achieves the highest DE (\(\approx\)0.97), followed by Prithvi-300M (\(\approx\)0.95), with U-Net++ and U-Net close behind (\(\approx\)0.94). In other words, across $\{10,2.5,1.25\}\%$ labels the Prithvi-600M model retains \(\approx\)97\% of its full-data performance on average about 2.9-3.2 points higher than the CNN baselines.

Finally, at $100\%$ labels, Prithvi-300M achieves the strongest absolute score (71.30 mIoU), followed by U-Net (70.44) and Prithvi-600M (70.41). Taken together, these results indicate that (i) EO-native pretraining confers clear advantages when labels are scarce, especially at the extreme $1.25\%$ (50 samples) regime; (ii) scaling Prithvi to 600M parameters improves robustness under scarcity even when full-data mIoU is comparable; and (iii) classical CNNs can approach GeoFMs once $\geq 10\%$ of labels (reaching the size of 400 samples) are available, but they are distinctly less reliable at ultra-low label fractions.

\subsection{Geographic Generalizability and Domain Transfer}

\begin{figure}[ht]
	\centering
	\includegraphics[width=0.9\linewidth]{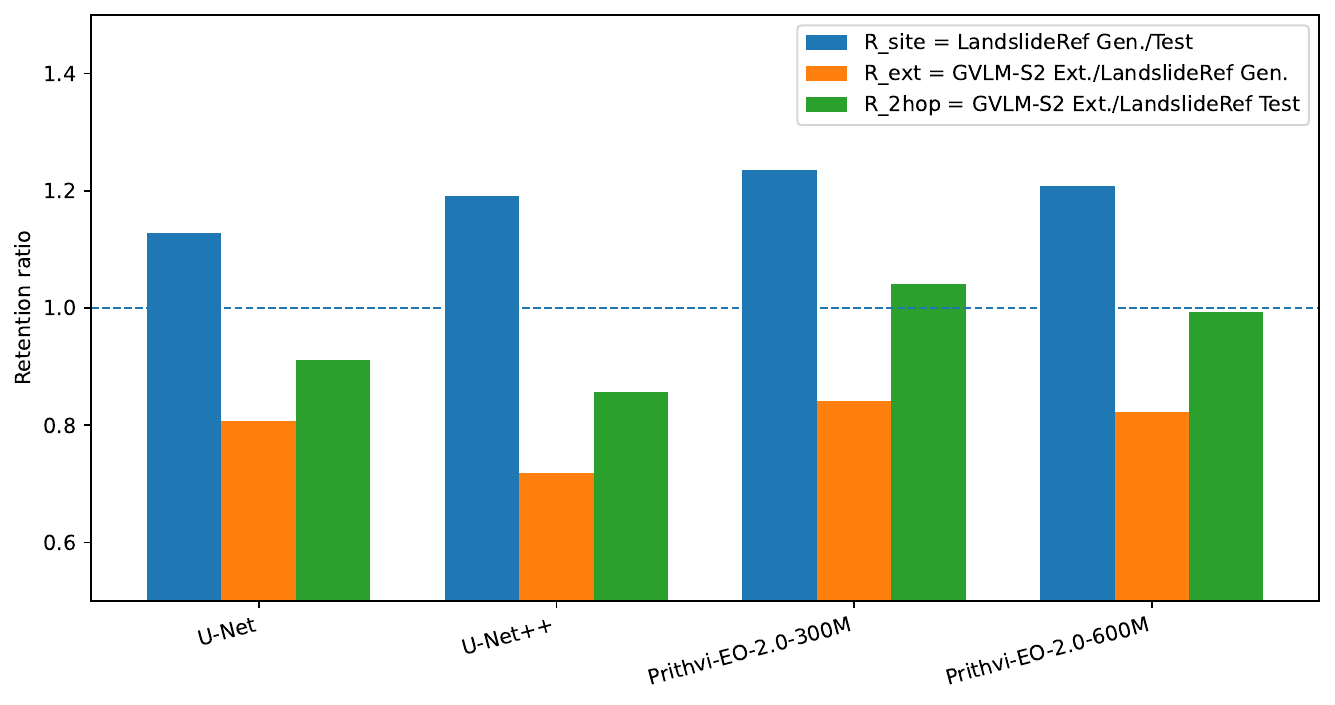} 
	\caption{Retention ratios summarizing transfer for each model. Blue bars: Gen/Test $=P(\mathcal{T}_{\mathrm{gen}})/P(\mathcal{T}_{\mathrm{in}})$ (within-corpus site shift). Orange bars: Ext/Gen $=P(\mathcal{T}_{\mathrm{ext}})/P(\mathcal{T}_{\mathrm{gen}})$ (cross-corpus shift normalized by the test-site difficulty). Green bars: Ext/Test $=P(\mathcal{T}_{\mathrm{ext}})/P(\mathcal{T}_{\mathrm{in}})$ (net change from test to external).}

	\label{fig:gen_retention}
\end{figure}

This section assessed model generalizability across domains, including both geographical regions and datasets. Two new datasets, Landslide Reference \citep{orynbaikyzy2025landslide} and GVLM-S2, were used because both provided location and timestamp information for the annotated landslides, and both used Sentinel-2 images as input to avoid cross-sensor data shifts. Table~\ref{tab:generalizability_table} and Figure~\ref{fig:gen_retention} together summarized cross-domain performance for both Prithvi-EO-2.0 and U-Net models. All models were trained using the training dataset of Landslide Reference and tested on three test sets: the testing set provided as part of the Landslide Reference data, the generalizability set provided in the same dataset, and the GVLM-S2 data, which we generated according to the labels in the original GVLM dataset and combined with the new Sentinel-2 image source. Both the generalizability test set and GVLM-S2 came from regions unseen in the Landslide Reference training data.

As Table~\ref{tab:generalizability_table} showed, on the Landslide Reference testing set, Prithvi-EO-2.0 consistently outperformed the U-Net family, with over a 10\% difference in mIoU (e.g., 71.18 for Prithvi-EO-2.0-600M vs. 58.23 for U-Net). Similar performance advantages of Prithvi-EO-2.0 were also observed on the other test datasets. These results again demonstrated the stronger generalizability of the GeoFM Prithvi-EO-2.0 compared to task-specific models such as U-Net.

\begin{table}[ht]
	\caption{Cross-domain evaluation on Landslide Reference (in-domain) and GVLM-S2 (external). All models are fine-tuned on the Landslide Reference training split (Sentinel-2 only; six HLS-like bands with B8 in place of B8A). Columns report mIoU on the official test split, the unseen-site generalizability split, and GVLM-S2.}

	\centering
    \begin{tabularx}{\textwidth}{ >{\raggedright\arraybackslash}X *{3}{>{\centering\arraybackslash}X}}
		\toprule
		\textbf{Model} & \textbf{Landslide Reference Testing Set} & \textbf{Landslide Reference Generalizability Set} & \textbf{GVLM-S2 \ \ \ \ \ \ \ \ \ \ \ External Set} \\
		\midrule
        U-Net (ResNet50) & 58.23 & 65.73 & 53.09 \\ 
        U-Net++ (ResNet50) & 56.71 & 67.59 & 48.61 \\
        Prithvi-EO-2.0-300M & 68.26 & 84.39 & \textbf{71.04} \\ 
        Prithvi-EO-2.0-600M & \textbf{71.18} & \textbf{86.03} & 70.75 \\ 
		\bottomrule
	\end{tabularx}
	\label{tab:generalizability_table}
\end{table}

When we cross-compared the performance on the Landslide Reference test set and its own generalizability test, the performance advantage of Prithvi-EO-2.0 was further shown. The mIoU gap between U-Net++ and Prithvi-EO-2.0-600M enlarged to over 20\% (67.59 vs. 86.03). This result demonstrated the stronger geographical generalizability of the Prithvi models, especially in unseen regions. This capability was extremely critical in real-world disaster monitoring situations.

An unusual pattern we observed in this result was that the model performance on the generalizability test set was consistently higher than that on the test set of the Landslide Reference data, which came from the same geographical regions as the training dataset. For example, Prithvi-EO-2.0-600M yielded 86.03 mIoU on the generalizability test and only 71.18 on the test set. This occurred because, in the test set, the Peat Landslides in Ireland performed unusually poorly, which depressed the aggregated test score \citep{orynbaikyzy2025landslide}. This reflects an important factor influencing model generalizability beyond geography. Although it is well recognized that location is a major factor affecting the geographical generalizability of deep learning models, in this case, landslide type played a more significant role than geography due to its morphological complexity and rare occurrences compared to other landslides. 

Furthermore, we observed that the Prithvi-EO-2.0 models transferred more reliably to the external GVLM-S2 than the U-Net baselines. Both variants sustained around 70 mIoU on GVLM-S2, while U-Net and U-Net++ dropped to the high-40s and low-50s. Within the Prithvi family, the 600M model attained the strongest in-domain generalizability score (on both test sets from the Landslide Reference data), whereas the 300M was marginally higher on the external dataset GVLM-S2. This suggested that model scaling (from 300M to 600M parameters) mainly strengthened robustness to within-corpus(dataset) variation, while both retained strong cross-corpus transfer.

The retention plot in Figure~\ref{fig:gen_retention} provides three complementary views of domain transfer. The green bars (R\_2hop) capture the net shift from the Landslide Reference test set to GVLM-S2. Here, the Prithvi models remain close to parity (ratios near 1.0), whereas the CNN baselines fall well below, indicating compounding sensitivity to both the challenging test site and the corpus shift. The orange bars (R\_ext) isolate the cross-corpus shift by normalizing out the test-set artifact, giving a clearer picture of domain robustness. On this view, Prithvi again stands well above the CNNs, maintaining its stability once the effect of the hard test site is discounted. Finally, the blue bars (R\_site) consistently exceed 1.0, confirming that the generalizability split is easier than the official test set and that the depressed test scores stem from site difficulty rather than an overall failure to generalize within the Landslide Reference dataset.

Overall, the table and retention figure convey a consistent message: (i) the Landslide Reference test split contains a hard site that distorts in-domain comparisons; (ii) after accounting for this via the R\_2hop view, Prithvi demonstrates substantially better cross-corpus stability than task-specific CNNs; and (iii) scaling Prithvi from 300M to 600M mainly strengthens within-corpus generalization while leaving external transfer strong for both variants.

\section{Discussion}
\begin{figure}[ht]
	\centering
	\includegraphics[width=0.7\linewidth]{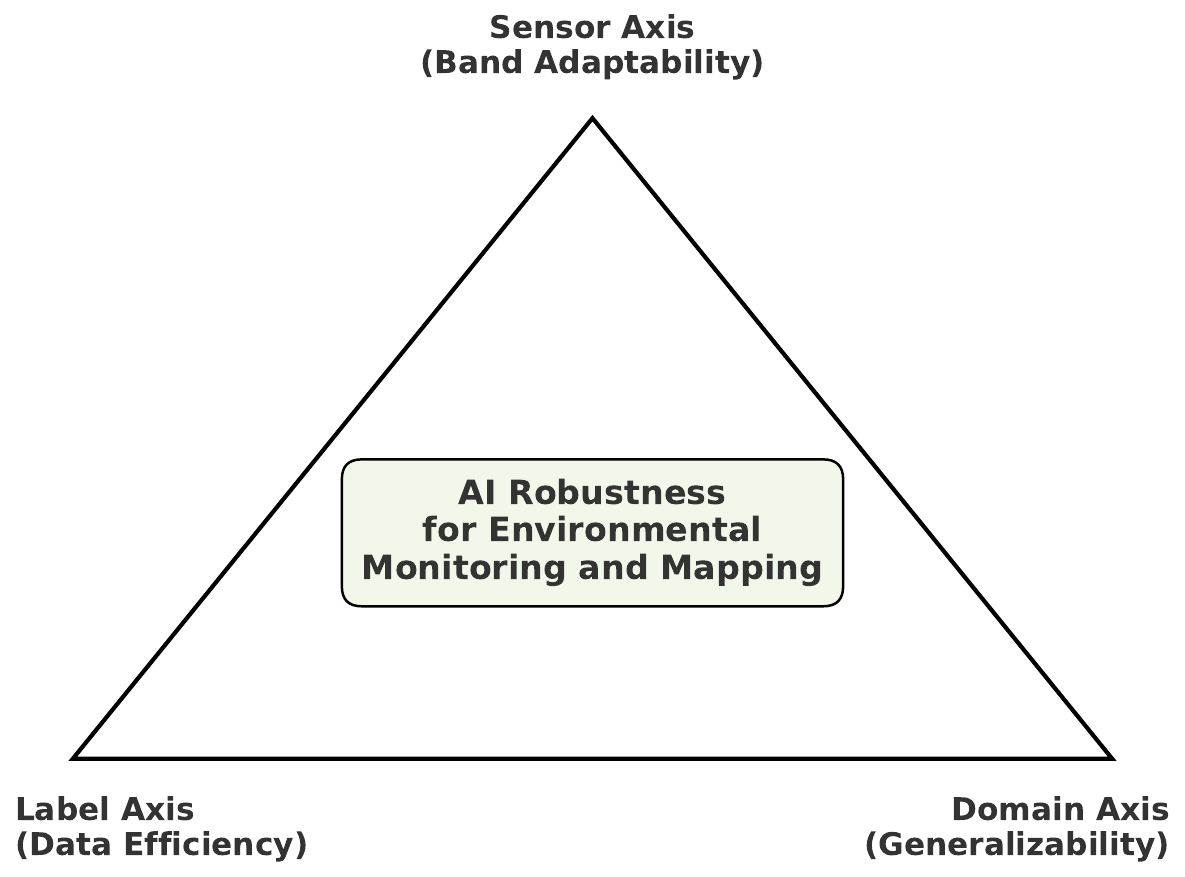} 
	\caption{A three-axis analytical framework for AI robustness in environmental monitoring and mapping}
	\label{fig:figure10}
\end{figure}

In this study, we examined the performance of geospatial foundation models for landslide mapping along three complementary axes - sensor, label, and domain that together capture the major sources of variability in real-world Earth observation workflows. The sensor axis reflects how well models adapt to different spectral configurations, the label axis evaluates robustness under limited supervision, and the domain axis tests generalizability across geography and datasets. This three-dimensional analytical framework (Figure~\ref{fig:figure10}) provides a structured lens for understanding the strengths and limitations of GeoFM in landslide mapping, and more broadly serves as a general framework for evaluating AI robustness in environmental monitoring and mapping.

First, our analysis addressed band adaptability as a key measure of GeoFM robustness, since in real-world applications the number and type of available bands often differ from the pretraining regime. Prithvi-EO-2.0, pretrained on EO-specific data, showed strong resilience to spectral changes, outperforming the popular U-Net models, the second best-performing model family on the Landslide4Sense training data. Nonlinear adaptation layers provided an effective compromise: frozen encoders with adapters performed well under limited resources, while full tuning delivered additional gains. This strategy also enabled the models to incorporate more than the six bands originally used by Prithvi, generating richer features and boosting predictive performance. We further found that band physics also mattered. Carefully selected six-band inputs aligned with physical drivers such as vegetation and soil moisture nearly matched the performance of richer inputs, whereas disrupting band identity or order caused sharp declines. Overall, effective deployment requires not just more bands, but thoughtful selection, consistent ordering, and, when needed, lightweight adaptation strategies to bridge training–operational gaps.

Second, the analysis along the label axis reiterates the value of pretraining for data efficiency, especially when annotated examples were extremely limited. In real-world disaster settings, where inventories are often incomplete, imbalanced, and slow to compile, Prithvi-EO-2.0 was far more reliable than the popular U-Net models with only a handful of labels. Even under severe scarcity, the pretrained model maintained useful accuracy, and scaling to larger parameter counts further improved robustness without substantially affecting full-data performance. Conventional CNNs narrowed the gap once moderate amounts of training data were available, but their instability at low label fractions made them less dependable for rapid-response mapping. These findings suggest that GeoFMs are particularly well suited for situations with scarce labels, supporting faster adaptation and more equitable deployment in time-critical disaster contexts.

Third, the analysis across domains showed that Prithvi-EO-2.0 generalized more effectively than task-specific CNNs when applied to unseen regions and independent datasets. It consistently outperformed U-Net models on the Landslide Reference benchmarks and maintained much stronger performance on the external GVLM-S2 dataset, where conventional baselines dropped sharply. The widening performance gap on the generalizability split highlighted Prithvi’s resilience to geographic shifts, a property especially important for operational disaster response. At the same time, the unexpected case of peat landslides in Ireland showed that domain transfer depends not only on spatial heterogeneity but also on event-specific factors, such as landslide type, which can outweigh geography. Scaling up Prithvi reinforced robustness within the training corpus, while both model sizes preserved strong cross-corpus transfer, demonstrating the value of larger encoders for stability without sacrificing adaptability to new datasets.

Despite its strong performance, Prithvi-EO-2.0 also has notable limitations. As a large model, it requires substantially more training time and computational resources than lighter architectures such as U-Net. Its complexity also demands greater technical expertise to adapt or extend, for instance when incorporating new components like band adapters. In addition, the model does not yet offer a complete task-specific workflow, e.g., for image segmentation, which has been proven effective in the model's predictive performance \citep{hsu2025geospatial}. While 74\% accuracy (refer to Table \ref {tab:prithvi_channel_adapt}) marks a clear improvement over baselines, the remaining 26\% misclassification rate still poses risks for disaster management, highlighting the need for fully developed operational pipelines. Another challenge lies in the limited reusability of existing AI-ready landslide training data. Although many datasets are available, they are often distributed as small tiles without standardized location or temporal metadata, which hampers reuse and location-based analysis of AI models. Exposing models to richer, domain-aware datasets would likely further enhance performance. To this end, future efforts should encourage data providers to release not only processed samples but also detailed metadata to enable AI-ready reuse and improve generalization.

\section{Conclusion}

This paper introduces a three-axis framework of sensor, label, and domain for evaluating the adaptability of geospatial foundation models in landslide mapping, and demonstrates how AI GeoFMs can strengthen disaster monitoring. Our experiments showed that Prithvi-EO-2.0, with EO-native pretraining and geospatial awareness, consistently outperformed task-specific CNNs, vision transformers, and other GeoFMs by achieving stronger band adaptability, data efficiency, and cross-domain generalizability. This study moves beyond simple performance evaluation to show how the core architectural principles of GeoFMs, including global pretraining, self-supervision, and adaptable fine-tuning, translate into scientific capabilities of flexibility, efficiency, and scalability. These qualities are essential for building proactive landslide risk management strategies and for addressing broader challenges in environmental monitoring and mapping.

In the future, we plan to strengthen the domain adaptation capabilities of GeoFMs by incorporating new strategies such as visual prompt tuning, which can reduce computational demands during fine-tuning and make the models more suitable for resource-constrained settings. We also aim to develop task-specific pipelines to further improve Prithvi’s performance across mapping and environmental analysis tasks. In addition, because disasters such as landslides often reshape landscapes, we will extend landslide mapping from relying solely on post-disaster imagery to integrating both pre- and post-disaster data, enabling more accurate quantification of landscape change caused by disasters. These strategies will enhance adaptability, lower computational barriers, and improve performance, thereby increasing GeoFMs’ reliability in operational settings. Such efforts will help unlock the full potential of GeoFMs, enabling robust AI-driven approaches not only for landslide risk reduction but also for environmental monitoring and sustainable development more broadly.

\bibliographystyle{unsrtnat}
\bibliography{references}  

\end{document}